\documentclass{zgroup/zgroup} 
\usepackage[utf8]{inputenc}
\usepackage{times}
\usepackage{url}

\usepackage{graphicx}
\usepackage{amsmath}
\usepackage{amssymb}
\usepackage{algorithm,algorithmic}
\usepackage{amsfonts,bm}
\usepackage{bbm}
\usepackage{eqparbox}

\usepackage{wrapfig}
\usepackage{url}
\usepackage{lipsum}
\usepackage{multicol}
\usepackage{multirow}
\usepackage{xcolor}
\usepackage{colortbl}
\usepackage{enumitem}
\usepackage{booktabs}
\usepackage[font=small]{caption}

\newif\ifarxiv
\arxivtrue
\newif\ifcomments
\commentstrue

\newcommand{\mr}[2]{\multirow{#1}{*}{#2}}%

\ifarxiv
\newcommand{\savespace}[1]{}
\newcommand{\savespaceeqn}{}
\newcommand{\savespacefig}{}
\newcommand{\savespacefigtop}[1]{}
\newcommand{\savespacebeforesection}{}
\newcommand{\savespacebeforeitem}{}
\else
\newcommand{\savespaceeqn}{\vskip -0.5cm}
\newcommand{\savespacefigtop}[1]{\vspace{#1}}
\newcommand{\savespace}[1]{\vspace{#1}}
\newcommand{\savespacefig}{}
\newcommand{\savespacebeforesection}{\vspace{-0.1in}}
\newcommand{\savespacebeforeitem}{\vspace{-0.03in}}
\fi

\newcommand{\gr}{\cellcolor[gray]{0.9}}
\newcommand{\gt}[1]{\textcolor{red}{#1}}
\newcommand{\sr}{\scriptsize}

\usepackage{xspace}

\newcommand*{\eg}{e.g.\@\xspace}
\newcommand*{\ie}{i.e.\@\xspace}

\makeatletter
\newcommand*{\etc}{%
    \@ifnextchar{.}%
        {etc}%
        {etc.\@\xspace}%
}
\makeatother
\newcommand{\mc}{\multicolumn}
\newcommand{\mrr}{\multirow}
\newcommand{\ul}{\underline}
\newcommand{\uftpn}{UFTE}
\newcommand{\uftsa}{UFTA}
\newcommand{\cm}{\checkmark}

\definecolor{red}{RGB}{255,73,92}
\definecolor{blue}{RGB}{37,110,255}
\definecolor{violet}{RGB}{70,35,122}
\definecolor{green}{RGB}{61,220,151}

\newcommand{\taskname}{FSAL}
\newcommand{\titlelower}{few-shot attribute learning}

\usepackage{amsmath,amsfonts,bm}

\def\eqref#1{equation~\ref{#1}}

\def\1{\bm{1}}

\def\rvv{{\mathbf{v}}}

\def\rvx{{\mathbf{x}}}

\newcommand{\bx}{\mathbf{x}}
\newcommand{\bz}{\mathbf{z}}
\newcommand{\bc}{\mathbf{c}}
\newcommand{\ba}{\mathbf{a}}
\newcommand{\bzeta}{\boldsymbol{\zeta}}

\newcommand{\bbR}{\mathbb{R}}

\DeclareMathAlphabet{\mathsfit}{\encodingdefault}{\sfdefault}{m}{sl}
\SetMathAlphabet{\mathsfit}{bold}{\encodingdefault}{\sfdefault}{bx}{n}

\def\gQ{{\mathcal{Q}}}

\def\gS{{\mathcal{S}}}

\newcommand{\sigmoid}{\sigma}

\DeclareMathOperator*{\argmin}{arg\,min}

\makeatletter
\def\thanks#1{\protected@xdef\@thanks{\@thanks
        \protect\footnotetext{#1}}}
\makeatother

\title[]{Probing Few-Shot Generalization with Attributes}

\zgroupauthor{\Name{Mengye Ren${}^*$}\thanks{*Equal contribution} \Email{mengye@nyu.edu}\\
  \addr New York University; Google\\
  \Name{Eleni Triantafillou${}^*$} \Email{etriantafillou@google.com}\\
  \addr Google\\
  \Name{Kuan-Chieh Wang${}^*$} \Email{wangkua1@stanford.edu}\\
  \addr Stanford University\\\
  \Name{James Lucas${}^*$} \Email{jlucas@cs.toronto.edu}\\
  \addr NVIDIA\\
  \Name{Jake Snell} \Email{jsnell@cs.toronto.edu}\\
  \addr University of Toronto; Vector Institute\\
  \Name{Xaq Pitkow} \Email{xaq@rice.edu}\\
  \addr Baylor College of Medicine; Rice University\\
  \Name{Andreas S. Tolias} \Email{astolias@bcm.edu}\\
  \addr Baylor College of Medicine; Rice University\\
  \Name{Richard Zemel} \Email{zemel@cs.columbia.edu}\\
  \addr Columbia University; University of Toronto; Vector Institute; 
  Canadian Institute for Advanced Research\\
}

\begin{document}

\maketitle

\vspace{-0.2in}
\begin{abstract}
    Despite impressive progress in deep learning, generalizing far beyond the
training distribution is an important open challenge. In this work, we consider
few-shot classification, and aim to shed light on what makes some novel classes
easier to learn than others, and what types of learned representations
generalize better. To this end, we define a new paradigm in terms of
\emph{attributes}---simple building blocks of which concepts are formed---as a
means of quantifying the degree of relatedness of different concepts. Our
empirical analysis reveals that supervised learning generalizes poorly to new
attributes, but a combination of self-supervised pretraining with supervised
finetuning leads to stronger generalization. The benefit of self-supervised
pretraining and supervised finetuning is further investigated through
controlled experiments using random splits of the attribute space, and we find
that predictability of test attributes provides an informative estimate of a
model's generalization ability.
\end{abstract}
\vspace{0.2in}
\section{Introduction}

While deep learning has led to numerous impressive success stories in recent
years, generalizing far beyond the training distribution is a lingering
challenge. Few-shot learning is a growing research area that aims at studying
and improving upon a model's ability to learn, for instance, new object
classes, from only a few examples. However, traditional few-shot learning
benchmarks are simplistic: while the test classes are disjoint from the
training classes, they often represent visually and semantically similar
concepts~\citep{lake2011oneshot,matchingnet}. Therefore it is difficult to
measure whether performance on these benchmarks is indicative of generalization
ability more broadly. Some recent benchmarks attempt to further separate train
and test classes, by splitting at a higher semantic level when a class
hierarchy is available \citep{fewshotssl} or holding out entire datasets
\cite{closerlook,broadstudy,triantafillou2019meta}, thus creating tougher
generalization problems, but we are still lacking a comprehensive study of what
underlies the ability to generalize better to some classes than to others.

\looseness=-10000
In this work, we study this question through the lens of representation
learning. We propose a new paradigm---\titlelower{} (\taskname{})--for probing
models' few-shot generalization ability, based on \emph{attributes}: simple
building blocks that can be used to define class concepts, \eg, \emph{birds}
are warm-blooded vertebrates that lay eggs and have feathers. Humans also
leverage similarity in the attribute space to recognize classes, which are
``information-rich bundles of attributes that form natural
discontinuities''~\citep{roschmervis1975family}. We use the relationship
between attributes and classes to design a framework to measure generalization
difficulty. Intuitively, if novel classes rely on attributes that were relevant
for training classes, albeit perhaps different combinations of them, then it
seems natural that those novel classes can be readily recognized with just a
few labeled examples. But what if novel classes rely on attributes that are
undefined or irrelevant during training? Will these classes be hard to learn?

Earlier empirical studies have examined the difficulty of few-shot learning
based on other notions of similarity that, for instance, relies on the WordNet
hierarchy \cite{sariyildiz2021}, or similarity of classes in the features space
of pre-trained models \cite{arnold2021embedding}. Compared to these works, we
directly leverage attributes to enable a more controlled study of
tranferability and few-shot generalization. Empirically, we also explore both
unsupervised and supervised approaches, revealing notably that a hybrid
self-supervised and supervised approach achieves stronger generalization
compared to other alternatives.

To summarize, our primary contributions are: 1) A new paradigm, FSAL, for
studying generalization in few-shot learning; 2) Three new datasets serving as
benchmarks for FSAL; 3) A study and analysis of different representation
learning methods and their generalization capabilities in these tasks.
\begin{figure*}[t]
\centering
\savespacefig
\savespacefig
\savespacefig
\savespacefig
\savespacefig
\includegraphics[width=0.97\linewidth,trim={0 0 0 0}, clip]{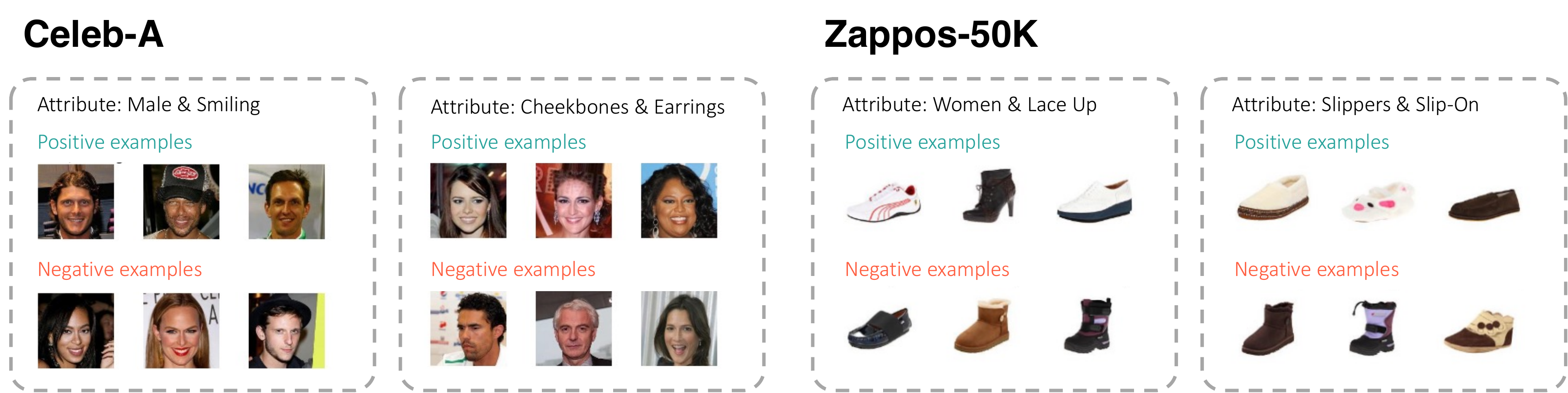}
\savespacebeforesection
\caption{\textbf{Sample \taskname{} episodes using Celeb-A (left) and Zappos-50K (right).}
Positive and negative examples are sampled according to attributes.}
\label{fig:sample}
\savespacebeforesection
\end{figure*}

\savespacebeforesection
\section{Few-Shot Attribute Learning}
\savespacebeforesection

In this section, we define our \titlelower{} (\taskname{}) paradigm and
highlight the additional challenges of \taskname{} compared to the standard
few-shot learning of semantic classes.

Similar to standard few-shot learning (FSL), at test time, the learner is presented
with an episode of data. The support set consists of $N$ positive and negative
examples of the target attributes
\ifarxiv
\begin{align}
\gS = \{ (\rvx^{S+}_1, 1), \dots, (\rvx^{S+}_N, 1), (\rvx^{S-}_1, 0), \dots,
(\rvx^{S-}_N, 0) \},
\end{align}
\else
$\gS = \{ (\rvx^{S+}_1, 1), \dots, (\rvx^{S+}_N, 1), (\rvx^{S-}_1, 0), \dots,
(\rvx^{S-}_N, 0) \},$
\fi
where the  $+$ or $-$ superscript suffix denotes whether the input is a
positive or negative example. 
After rapid learning on the support set, the
model is then evaluated on the binary classification performance of the query
set:
\ifarxiv
\begin{align}
\gQ = \{ (\rvx^{Q+}_1, 1), \dots, (\rvx^{Q+}_M, 1), (\rvx^{Q-}_1, 0), \dots,
(\rvx^{Q-}_M, 0) \}.
\end{align}
\else
$\gQ = \{ (\rvx^{Q+}_1, 1), \dots, (\rvx^{Q+}_M, 1), (\rvx^{Q-}_1, 0), \dots,
(\rvx^{Q-}_M, 0) \}.$
\fi

\looseness=-1000
As is standard in FSL, before the test episodes, 
we allow methods to learn a representation.
In \taskname, this involves a labeled set of training attributes, but these must be disjoint from test attributes. For example, the model can learn attributes such as hair color and mustache during training, and will be tested on eyeglasses at test time.
Similar to standard representation learning in FSL,
training labels can be presented in the form of \textit{episodic labels} for meta-learning methods, or \textit{absolute labels} for pretraining-based methods. In \taskname{}, episodic labels refer to binary attribute labels in each episode, and absolute labels refer to attribute IDs. 

At test time, the target binary label may concern a novel attribute that was
previously unlabeled in the training set. For example, in one test episode, a
smiling face with \emph{eyeglasses} is positively labeled alongside other faces
with eyeglasses. The task here is to learn the attribute of ``wearing
eyeglasses''. However, while the learner might have seen training images with
eyeglasses, it was never a relevant feature for the purpose of predicting the
positively labeled instances. For simplicity, each test episode is a binary classification problem. It can be easily extended to multiple new attributes by considering a few binary classification problems at the same time.

Furthermore, suppose that in another test episode, the same \emph{smiling} face
is positively labeled alongside other smiling faces. The target attribute here
has now changed from ``wearing eyeglasses'' to ``smiling.'' This highlights a
critical difference between few-shot attribute learning and standard few-shot
learning of semantic classes: in standard FSL, each instance can belong to only
one class regardless of the episode. In FSAL, due to the multi-label nature of
the attribute space, one instance could have different labels depending on the
context of the support set examples. Furthermore, there may be a large amount
of ambiguity when the support set is small. Figure~\ref{fig:sample} shows a few
examples of our attribute learning episodes. Note that in order to create task
diversity, we allow both unary and binary attributes, where binary attributes
are conjunctions of two unary attributes.

In order to solve the \taskname{} task, the learner must correctly determine
the context. Just like in zero-shot learning, one natural way to solve this
problem would be to learn to predict the underlying attributes of each image.
Given the attributes, you could then estimate the context in each
episode~\citep{attributezsl}. However, methods that accurately predict
attributes relevant to training episodes may not generalize well, since at test
time \taskname{} introduces novel attributes. Instead, we explore methods that
allow more general representations to be learned.


\savespacebeforesection
\section{Experiment Methodology}
\savespacebeforesection
\label{sec:method}

In this section, we describe a range of methods that can be used for the
problem of
\titlelower{}. The methods can be organized into two stages. The first stage is
representation learning through either pre-training the network or performing
meta-learning. The second stage is learning a few-shot classifier at test-time
to solve a new episode. We describe each stage of learning below.

\savespacebeforesection
\subsection{Stage I: Representation Learning}
\label{sec:ft}
\savespacebeforesection
We consider the following representation approaches in our evaluation.
\savespacebeforesection
\paragraph{Supervised:} Many of the existing few-shot learning approaches
include a stage of supervised representation learning. Two classes of
approaches are frequently employed:
\begin{itemize}[leftmargin=*]
\savespacebeforesection
\item Episodic meta-learning approaches train directly from a set of few-shot
episodes using episodic labels. This class of methods can be naturally applied
to our learning setting.
\item Supervised classification approaches train a network to directly classify
a set of training classes using absolute labels, and at test time, the
embedding network is transferred to solve the test task by training another
classifier on top. If absolute attributes are provided to the learner, then one
natural approach is to instead train an attribute classifier with multiple
binary outputs. After the attribute classifier network is learned, we can then
transfer the representations to recognize test attributes. We denote this
method as Supervised Attributes (\textbf{SA}).
\end{itemize}
\savespacebeforesection
\paragraph{Unsupervised:} 
As supervised representation learning may not generalize to novel attributes,
we also consider unsupervised representation learning as another option. We
chose SimCLR~\citep{simclr} as a representative from this category due to its
empirical success. In general, contrastive learning approaches aim to build
invariant representations between a pair of inputs $\{\mathbf{x},
\mathbf{x'}\}$ that are produced by applying random data augmentations (e.g.
cropping) to an input image. It is likely to preserve more general semantic
features since all attributes are useful towards identifying another random
crop of the same image. We first obtain the embedding output $\mathbf{h}$ from
the CNN, and then following \citep{simclr}, we project $\mathbf{h}$ to
$\mathbf{z}$ using a multi-layered perceptron (MLP): $\mathbf{h} =
\mathrm{CNN}(\mathbf{x}), \mathbf{z} =
\mathrm{MLP}_1(\mathbf{h})$. With a batch of image pairs denoted by
$\{\mathbf{x}_i\}, \{ \mathbf{x}_i'\}$, we can obtain their features
$\{\mathbf{z}_i\}, \{ \mathbf{z}_i'\}$, and the contrastive loss function is
defined similar to the cross entropy function:
\ifarxiv
\begin{align}
    \mathcal{L}_1 = -\sum_i \log\frac{\exp(\mathbf{z}_i \cdot \mathbf{z}_i' /
    \tau)}{\sum_j
    \exp(\mathbf{z}_i \cdot \mathbf{z}_j' / \tau)},
\end{align}
\savespaceeqn
\else
$
    \mathcal{L}_1 = -\sum_i \log\frac{\exp(\mathbf{z}_i \cdot \mathbf{z}_i' /
    \tau)}{\sum_j
    \exp(\mathbf{z}_i \cdot \mathbf{z}_j' / \tau)},
$
\fi
where $\tau$ is a temperature parameter. We denote Unsupervised representation
learning as \textbf{U}.

\savespacebeforesection
\paragraph{Unsupervised-then-Finetuning:} For unsupervised learning, we also
consider adding a subsequent stage of supervised fine-tuning to utilize
attribute labels from the training set. Note that fine-tuning here is different
from fine-tuning in regular few-shot learning as it is not fine-tuning on test
episodes but rather on the original training set. To prevent overwriting the
representations and making them overly sensitive to training attributes, we add
another projection MLP that learns more specific representations for finetuning
on training attributes: $\mathbf{g} = \mathrm{MLP}_2(\mathbf{h}).$ Here, we
again consider using two different modes of supervision: 1) the \taskname{}
binary episodic labels, or 2) the underlying absolute attribute labels:
\begin{itemize}[leftmargin=*]
\savespacebeforesection
\item Unsupervised-then-FineTune-on-Episodes (\textbf{\uftpn}).
We adopt the
Prototypical Networks~\citep{protonet} formulation, where the network solves a
learning episode of $N$ positive and negative support examples by using
prototypes $\mathbf{p}$: $\mathbf{p}^+ = \frac{1}{N} \sum_i \mathbf{g}^+_i;
\mathbf{p}^- =
\frac{1}{N} \sum_i \mathbf{g}^-_i.$ With query example $\mathbf{g}^q$, we can
make a binary prediction:
$\hat{y}^q = \frac{\exp(-d(\mathbf{g}^q,
    \mathbf{p}^+))}{\exp(-d(\mathbf{g}^q, \mathbf{p}^+)) +
    \exp(-d(\mathbf{g}^q, \mathbf{p}^-))},$
where $d$ is some dissimilarity score, \eg Euclidean distance or cosine
dissimilarity, and the training objective is to minimize the classification
loss between the prediction $\hat{y}^q$ and the label $y^q$:
\ifarxiv
\begin{align}
\mathcal{L}_{2E} &= \sum_j - y_j \log \hat{y}^q_j - (1-y^q_j) \log (1 -
\hat{y}^q_j),
\end{align}
where $j$ is the index of query examples.
\else
$\mathcal{L}_{2E} = \sum_j - y_j \log \hat{y}^q_j - (1-y^q_j) \log (1 -
\hat{y}^q_j),$
where $j$ is the index of query examples.
\fi
    
\item Unsupervised-then-FineTune-on-Attributes (\textbf{\uftsa}). With
persistent attribute information, we can train a linear classifier with sigmoid
activation to directly predict the absolute attribute labels $\mathbf{a}$:
$\hat{\mathbf{a}} = W_A \mathbf{g} + b_A$, with the loss being
\ifarxiv
\begin{align}
\mathcal{L}_{2A} &= \sum_k - \mathbf{a}_k \log \hat{\mathbf{a}}_k -
(1-\mathbf{a}_k) \log (1 -
\hat{\mathbf{a}}_k),
\end{align}
\savespaceeqn
\vskip -0.6cm
where $k$ is the index of attributes.
\else
$\mathcal{L}_{2A} = \sum_k - \mathbf{a}_k \log \hat{\mathbf{a}}_k -
(1-\mathbf{a}_k) \log (1 -
\hat{\mathbf{a}}_k),$
where $k$ is the index of attributes.
\fi
\end{itemize}

\savespacebeforesection
\subsection{Stage II: Few-Shot Learning}
\label{sec:fsl}
\savespacebeforesection

Once representations are learned, it remains to be decided how to use the small
support set of each given test episode in order to make predictions for the
associated query set. For each model described in the previous stages, we
consider three candidate approaches: nearest neighbor (\textbf{NN}) used in
MatchingNet~\citep{matchingnet}, the nearest centroid (\textbf{NC}) used in
ProtoNet~\citep{protonet}, and logistic regression (\textbf{LR}) used in
\citet{closerlook}. The LR approach learns a weight coefficient for each
feature dimension, thus performing some level of feature selection, unlike the
NC or NN alternatives. In addition, we apply an L1 regularizer on LR to
encourage sparsity. In this way, the learning of a classifier is essentially
done at the same time as the selection of feature dimensions. The overall
objective of the classifier is:
\ifarxiv
\begin{align}
\argmin_{\mathbf{w}, b} - y \log(\hat{y}) - (1-y) \log(1 - \hat{y}) + 
\lambda \lVert \mathbf{w}
\rVert_1,
\end{align}
\else
$
\argmin_{\mathbf{w}, b} - y \log(\hat{y}) - (1-y) \log(1 - \hat{y}) + 
\lambda \lVert \mathbf{w}
\rVert_1,$
\fi
where $\hat{y} = \sigmoid(\mathbf{w}^\top \mathbf{h} + b)$, and $\mathbf{h}$ is
the representation vector extracted from the CNN backbone. Note that in this
stage we discard the projection MLPs that are defined in previous stages since
they are trained towards training attributes and self-supervised objectives,
and we found that they do not transfer well to novel attributes.
\begin{table*}
\savespacefig
\savespacefig
\savespacefig
\savespacefig
\begin{center}
\begin{small}
\ifarxiv
\resizebox{\textwidth}{!}{
\begin{tabular}{cll}
\toprule
\bf Paradigm                  & \bf Test time task                                   & \bf Task specification     \\
\hline                                                                                                                  ZSL~\citep{attributezsl}      & Novel semantic classes of existing attributes        & Attribute IDs         \\
CZSL~\citep{czsl}             & Novel combinations of existing attributes \& classes & Attribute IDs \\
FSL~\citep{lake2011oneshot}   & Novel semantic classes                        & Support examples           \\   
\midrule
\taskname{} (Ours)             & Novel (previously unlabeled) attributes      & Support examples           \\
\bottomrule
\end{tabular}
}
\else
\resizebox{0.8\textwidth}{!}{
\begin{tabular}{cll}
\toprule
\bf Paradigm                  & \bf Test time task                                   & \bf Task specification     \\
\hline                       
ZSL~\citep{attributezsl}      & Novel semantic classes of existing attributes        & Attribute IDs         \\
CZSL~\citep{czsl}             & Novel combinations of existing attributes \& classes & Attribute IDs \\
FSL~\citep{lake2011oneshot}   & Novel semantic classes                        & Support examples           \\   
\midrule
\taskname{} (Ours)             & Novel (previously unlabeled) attributes      & Support examples           \\
\bottomrule
\end{tabular}
}
\fi
\end{small}
\end{center}
\savespacebeforesection
\caption{\textbf{Differences between  zero-shot learning (ZSL), compositional ZSL (CZSL), few-shot
learning (FSL), and our newly proposed \titlelower{} (\taskname{}).} Our task requires the model to
generalize to new attributes.}
\label{tab:benchmarkdiff}
\savespacebeforesection
\end{table*}
\savespacebeforesection
\section{Related Work}
\savespacebeforesection
\paragraph{Few-shot learning:}
Few-shot learning (FSL)~\citep{fei2006one,lake2011oneshot,matchingnet} entails
learning new tasks with only a few examples. With an abundance of training
data, FSL is closely related to the general meta-learning or learning to learn
paradigm~\citep {Thrun1998}, as a few-shot learning algorithm can be developed
on training tasks and run on novel tasks at test time. In standard few-shot
classification, each image only has a single unambiguous class label, whereas
in our few-shot attribute learning, the target attributes can vary depending on
how the support set is presented. We show in this paper that this is a more
challenging problem as it requires the model to be more flexible and
generalizable. In early benchmarks, a set of semantic classes was randomly
split into a training and test set. We hypothesize that this often leads to a
common set of attributes that span (most of) the training and test classes,
thus causing high transferability between these two sets, which allows simple
solutions based on feature re-use \citep{closerlook,anil} to work well. Later
benchmarks explicitly attempt to vary the separation between train and test
classes, based on varying the distances in the underlying WordNet classes
(\textit{tiered}-ImageNet \citep{fewshotssl}), or in different image domains
(Meta-Dataset \citep{triantafillou2019meta}). However, we argue that reasoning
about the underlying attributes directly offers a more systematic framework to
measure the relatedness and transferability between the train and test set. We
expect our analysis to open the door to such studies in the future. Few-shot
attribute learning is also related to multi-label few-shot
learning~\citep{alfassy2019laso,li2021compositional}. These prior works
emphasize on the compositional aspect, whereas we propose models that address
the transferability of the learned representations. Additionally,
\citet{xiang2019incremental} explored combining incremental few-shot learning
and attribute learning for pedestrian images.

\savespacebeforesection
\paragraph{Attribute learning:}
In the past, there have been a number of works that aim to predict attribute
information from raw
inputs~\citep{ferrari2007attribute,farhadi2009describe,farhadi2010attribute,wang2010discriminative}.
A related model is later proposed by \citet{koh2020concept} to achieve better
causal interpretability.
There have also been a number of datasets that have been collected with visual
attributes annotated
\citep{cub,zappos,celeba,patterson2016coco,pham2021attribute}. One key
difference between our work and these attribute learning approaches is that at
test time we aim to learn a classifier on novel attributes that are previously
not labeled in the training set, and this brings additional challenges of
transfer learning and learning with limited labeled data.

\savespacebeforesection
\paragraph{Zero-shot learning:} In zero-shot learning
(ZSL)~\citep{farhadi2009describe,labelembed,goodbadugly,attributezsl,ezzsl,evaluateoutput},
a model is asked to recognize classes not present in the training set,
supervised only by some auxiliary description~\citep{descriptionzsl} or
attribute values~\citep{farhadi2009describe} (see~\citet{wang2019survey} for a
survey). \citet{attributezsl} studied the \textit{direct attribute prediction}
method, similar to the Supervised Attribute baseline described in
Section~\ref{sec:baselines}. Compositional ZSL aims at learning
classes~\citep{czsl,taskdriven,taskaware,unseencomposition} defined by a novel
composition of labeled attributes and object classes. An important distinction
between ZSL and our few-shot attribute learning task is that ZSL uses the same
set of attributes for both training and testing; by contrast, our task asks the
model to learn attributes for which there are no labels during training, and
they may not be relevant to any of the training attributes or episodes. We
summarize the relationships between ZSL, FSL and our task in
Table~\ref{tab:benchmarkdiff}.


\savespacebeforesection
\paragraph{Generalization to novel tasks:}
\looseness=-1 
One key component of our work is an attempt to understand the generalization
behavior of learning novel concepts at test time. Relevant theoretical studies
consider novel task generalization, casting it in a transfer learning and
learning to learn
framework~\citep{baxter2000model,ben2008notion,ben2010theory,pentina2014,amit2018meta,lucas2020lb}.
A common theme in these studies is in characterizing task relatedness, and the
role that it plays in generalization to novel tasks.
\citet{arnold2021embedding} studied task clustering for few-shot learning in
the embedding space and found class splits that are of different difficulty
levels. \citet{sariyildiz2021} use the WordNet hierarchy to compute semantic
distances. In our paper, we instead split the data in the attribute space, and
if we assume that semantic classes are combinations of attributes, then a
disjoint attribute split will imply further semantic distances.  In our work,
we investigate the role of task relatedness empirically by investigating
generalization performance under different splits of the attribute space.
\begin{table}[t]
\savespacefigtop{-0.5in}
\begin{center}
\begin{small}

\resizebox{0.7\textwidth}{!}{
\begin{tabular}{lc|cc|cc}
\toprule
                     &           & \mc{2}{c|}{\bf Celeb-A}                                       & \mc{2}{c}{\bf Zappos-50K}                                      \\
                     &    Sup.   & 5-shot                        & 20-shot                       & 5-shot                        & 20-shot                        \\
\hline
Chance               &   -       & 50.00{\sr$\pm$0.00}           & 50.00{\sr$\pm$0.00}           & 50.00{\sr$\pm$0.00}           & 50.00{\sr$\pm$0.00}            \\
\hline
MatchingNet          & \it E     & 68.30{\sr$\pm$0.76}           & 71.73{\sr$\pm$0.52}           & 77.26{\sr$\pm$0.60}           & 80.47{\sr$\pm$0.49}            \\
MAML/ANIL            & \it E     & 71.24{\sr$\pm$0.74}           & 73.35{\sr$\pm$0.53}           & 77.05{\sr$\pm$0.50}           & 81.10{\sr$\pm$0.43}            \\
TAFENet              & \it E     & 69.10{\sr$\pm$0.76}           & 72.11{\sr$\pm$0.54}           & 79.20{\sr$\pm$0.57}           & 83.42{\sr$\pm$0.44}            \\
ProtoNet             & \it E     & 72.12{\sr$\pm$0.75}           & 75.27{\sr$\pm$0.51}           & 77.22{\sr$\pm$0.51}           & 83.42{\sr$\pm$0.41}            \\
TADAM                & \it E     & 73.54{\sr$\pm$0.70}           & 76.06{\sr$\pm$0.53}           & 81.45{\sr$\pm$0.50}           & 86.23{\sr$\pm$0.40}            \\
ID                   & \it C     & 69.95{\sr$\pm$0.69}           & 77.53{\sr$\pm$0.53}           & -                             & -                              \\
SA                   & \it A     & 72.91{\sr$\pm$0.74}           & 78.86{\sr$\pm$0.48}           & 82.17{\sr$\pm$0.48}           & 88.24{\sr$\pm$0.37}            \\
U                    & -         & 73.47{\sr$\pm$0.68}           & 79.97{\sr$\pm$0.51}           & 83.88{\sr$\pm$0.44}           & 90.92{\sr$\pm$0.32}            \\
\uftpn{}  & \it E & \ul{76.69}{\sr$\pm$0.69}  & \ul{82.83}{\sr$\pm$0.48}  & {\bf 85.50}{\sr$\pm$0.42} & {\bf 92.20}{\sr$\pm$0.28}  \\
\uftsa{}  & \it A & {\bf 78.98}{\sr$\pm$0.69} & {\bf 84.14}{\sr$\pm$0.48} & \ul{84.61}{\sr$\pm$0.43}  & {\bf 91.66}{\sr$\pm$0.29}  \\
\hline
                                                                                                                                                                  \\
\mc{6}{l}{\bf Oracles}                                                                                                                                            \\
\hline
SA*                  & \it A*     & 84.74{\sr$\pm$0.60}           & 89.15{\sr$\pm$0.38}           & 88.11{\sr$\pm$0.39}           & 93.00{\sr$\pm$0.28}            \\
GT                   & -         & 91.07{\sr$\pm$0.49}           & 98.16{\sr$\pm$0.17}           & 97.66{\sr$\pm$0.16}           & 99.84{\sr$\pm$0.04}            \\
\bottomrule
\end{tabular}
}
\end{small}
\end{center}
\caption{\textbf{5- and 20-shot attribute learning results on Celeb-A and
Zappos-50K.} 
Methods can be supervised by 1) \textit{``E''}=episode binary labels,
2) \textit{``A''}=attributes, and 3) \textit{``C''}=face identity. The best is
\textbf{bolded} and the second best is \ul{underlined}. }
\label{tab:main}
\savespacebeforesection
\savespacebeforesection
\end{table}

\savespacebeforesection
\section{Experiments}
\savespacebeforesection
In this section, we evaluate different representation and few-shot learning
approaches on \taskname{} using several datasets. 

\savespacebeforesection
\subsection{Datasets}
\savespacebeforesection
We consider the following three datasets:

\begin{itemize}[leftmargin=*]
\savespacebeforeitem
\item \textbf{Celeb-A}~\citep{celeba} contains over 200K images of faces. Each image is annotated with binary attributes, detailing hair color,
facial expressions, etc. We split 14 attributes for training and 13 for test.

\item \textbf{Zappos-50K}~\citep{zappos} contains just under 50K images of
shoes annotated with attribute values. We split these into 40 attribute values
for training, and 39 for testing.

\item \textbf{ImageNet-with-Attributes} is a small subset of the ImageNet
dataset~\citep{deng2009imagenet} with attribute annotations. It has
9.6K images. We used 11 attributes for training and 10 for testing. Note that this subset of ImageNet that has attribute labels is significantly smaller than the two datasets above, and it is not sufficiently large for meta-learning methods from scratch.  Hence, the results for this dataset are reported separately.

\end{itemize}
\savespacebeforesection
In all of the datasets above, there is no overlap between training and test
attributes. Additional split details can be found in the supplementary
materials.

\savespacebeforesection
\paragraph{Episode construction:} For each episode, we randomly select one or
two attributes and look for positive examples belonging to these attributes
simultaneously. We also sample an equal number of negative examples that don't
match the selected attributes. This will construct a \textit{support set} of
positive and negative samples, and then we repeat the same process for the
corresponding \textit{query set} as well. Sample episodes are shown in
Figure~\ref{fig:sample}.
Additional episodes are shown in the Appendix.

\begin{table}[t]
\begin{center}
\begin{small}
\resizebox{0.8\columnwidth}{!}{
\begin{tabular}{l|ccc|ccc}
\toprule
              & \mc{3}{c|}{\bf Celeb-A}                                                           & \mc{3}{c}{\bf Zappos-50K}                                                          \\
              & NN                      & NC                      & LR                            & NN                       & NC                       & LR                           \\
\hline                                                                                                                                                                             
Meta          & 71.73{\sr$\pm$0.52}     & 75.27{\sr$\pm$0.51}     & 73.38{\sr$\pm$0.53}           & 80.47{\sr$\pm$0.49}      & 83.42{\sr$\pm$0.41}      & 81.10{\sr$\pm$0.43}          \\
SA            & 75.33{\sr$\pm$0.47}     & 77.24{\sr$\pm$0.51}     & 78.86{\sr$\pm$0.48}           & 81.17{\sr$\pm$0.44}      & 85.48{\sr$\pm$0.41}      & 88.24{\sr$\pm$0.37}          \\
U             & 75.72{\sr$\pm$0.48}     & 77.78{\sr$\pm$0.52}     & 79.97{\sr$\pm$0.51}           & 85.17{\sr$\pm$0.40}      & 88.63{\sr$\pm$0.37}      & 90.92{\sr$\pm$0.32}          \\
\uftpn    & 79.03{\sr$\pm$0.47} & 81.04{\sr$\pm$0.47} & \ul{82.83}{\sr$\pm$0.48}  & 86.23{\sr$\pm$0.34}  & 90.61{\sr$\pm$0.31} & {\bf92.20}{\sr$\pm$0.28} \\
\uftsa    & 77.30{\sr$\pm$0.52} & 82.16{\sr$\pm$0.46} & {\bf 84.14}{\sr$\pm$0.48} & 86.40 {\sr$\pm$0.36}  & 90.25{\sr$\pm$0.33}  & {\bf91.66}{\sr$\pm$0.29} \\
\hline
SA*           & 78.84{\sr$\pm$0.41}     & 84.61{\sr$\pm$0.42}     & 89.15{\sr$\pm$0.38}           & 87.54{\sr$\pm$0.33}      & 90.97{\sr$\pm$0.31}      & 93.00{\sr$\pm$0.28}      \\
\bottomrule
\end{tabular}
}
\end{small}
\end{center}
\caption{\textbf{Combination of different representation \& few-shot learners
on 20-shot attribute learning.} 
Note: Meta-NN =
MatchingNet, Meta-NC = ProtoNet, Meta-LR = MAML/ANIL.}
\label{tab:combo}
\savespacebeforesection
\end{table}

\begin{table}[t]
\begin{center}
\begin{small}
\begin{minipage}[b]{0.52\linewidth}
\begin{center}
\ifarxiv
\resizebox{!}{1.2cm}{
\begin{tabular}{l|ccc|ccc}
\toprule
             & \mc{3}{c|}{\bf Celeb-A}                                               & \mc{3}{c}{\bf Zappos-50K}        \\
             & Train attr          & Test attr                 & Gap                 & Train attr          & Test attr           & Gap   \\
\hline                                                                                                    
ProtoNet     & 87.12{\sr$\pm$0.40} & 75.09{\sr$\pm$0.52}       & \color{red}{--12.03} & 92.88{\sr$\pm$0.24} & 83.42{\sr$\pm$0.41} &  \color{red}{--9.46}   \\
SA           & 88.25{\sr$\pm$0.38} & 78.86{\sr$\pm$0.48}       & \color{red}{--9.39}  & 95.11{\sr$\pm$0.19}    & 88.24{\sr$\pm$0.37} &  \color{red}{--6.87}   \\
U            & 79.48{\sr$\pm$0.54} & 79.97{\sr$\pm$0.51}       & --0.49               & 94.03{\sr$\pm$0.23} & 90.92{\sr$\pm$0.32} & {--3.11}    \\
\uftpn   & 87.25{\sr$\pm$0.40} & \ul{82.83}{\sr$\pm$0.48}  & --4.42               & 95.91{\sr$\pm$0.18} & {\bf 92.20}{\sr$\pm$0.28}  &      --3.71\\
\uftsa   & 85.53{\sr$\pm$0.43} & {\bf 84.14}{\sr$\pm$0.48} & --1.39               & 94.61{\sr$\pm$0.21} & {\bf91.66}{\sr$\pm$0.29} & --2.95 \\
\hline                                                                                                                                                    
SA*          & 87.88{\sr$\pm$0.39} & 89.15{\sr$\pm$0.38}       & +1.27               & 95.59{\sr$\pm$0.18} & 93.00{\sr$\pm$0.28} & --2.58\\
\bottomrule
\end{tabular}
}
\else
\resizebox{!}{1cm}{
\begin{tabular}{l|ccc|ccc}
\toprule
             & \mc{3}{c|}{\bf Celeb-A}                                               & \mc{3}{c}{\bf Zappos-50K}        \\
             & Train attr          & Test attr                 & Gap                 & Train attr          & Test attr           & Gap   \\
\hline                                                                                                    
ProtoNet     & 87.12{\sr$\pm$0.40} & 75.09{\sr$\pm$0.52}       & \color{red}{--12.03} & 92.88{\sr$\pm$0.24} & 83.42{\sr$\pm$0.41} &  \color{red}{--9.46}   \\
SA           & 88.25{\sr$\pm$0.38} & 78.86{\sr$\pm$0.48}       & \color{red}{--9.39}  & 95.11{\sr$\pm$0.19}    & 88.24{\sr$\pm$0.37} &  \color{red}{--6.87}   \\
U            & 79.48{\sr$\pm$0.54} & 79.97{\sr$\pm$0.51}       & --0.49               & 94.03{\sr$\pm$0.23} & 90.92{\sr$\pm$0.32} & {--3.11}    \\
\uftpn   & 87.25{\sr$\pm$0.40} & \ul{82.83}{\sr$\pm$0.48}  & --4.42               & 95.91{\sr$\pm$0.18} & {\bf 92.20}{\sr$\pm$0.28}  &      --3.71\\
\uftsa   & 85.53{\sr$\pm$0.43} & {\bf 84.14}{\sr$\pm$0.48} & --1.39               & 94.61{\sr$\pm$0.21} & {\bf91.66}{\sr$\pm$0.29} & --2.95 \\
\hline                                                                                                                                                    
SA*          & 87.88{\sr$\pm$0.39} & 89.15{\sr$\pm$0.38}       & +1.27               & 95.59{\sr$\pm$0.18} & 93.00{\sr$\pm$0.28} & --2.58\\
\bottomrule
\end{tabular}
}
\fi
\caption{
\textbf{ Comparison of representation learning methods with respect to their
ability to predict training and testing attributes.} Standard methods such as
ProtoNet and SA perform well on training attributes but do not transfer well to
novel ones (large training vs. test gaps in \textcolor{red}{red}).
}
\label{tab:gap}
\end{center}
\end{minipage}
\hfill
\begin{minipage}[b]{.46\linewidth}
\begin{center}
\ifarxiv
\resizebox{!}{1.3cm}{
\begin{tabular}{cccccccc}
\toprule
\mrr{2}{*}{L}& \mrr{2}{*}{D?}    &         \mc{2}{c}{\bf \uftpn}                   &  \mc{2}{c}{\bf \uftsa}                        \\
              &                                & Val Acc. ($\Delta$)      & Gap                  & Val Acc.  ($\Delta$)    & Gap                 \\
\hline                                                                                                                                      
0             &                                & 78.02 (--2.19)           & \color{red}{--9.72}  & 82.81 (+2.60)          & --4.80               \\
1             &                                & 76.86 (--3.35)           & \color{red}{--11.14} & 79.56 (--0.65)          & \color{red}{--7.43} \\
1         & \cm                        & \ul{82.01} (+1.80)   & --5.63           & \ul{83.39} (+3.18)  & --2.05          \\
2             &                                & 76.32 (--3.89)           & \color{red}{--11.58} & 79.71 (--0.50)          & \color{red}{--7.23} \\
2         & \cm                        & {\bf82.43} (+2.22)   & --4.83           & {\bf 83.86} (+3.65) & --1.90          \\
\bottomrule
\end{tabular}
}
\vspace{0.1in}
\else
\resizebox{!}{1.1cm}{
\begin{tabular}{cccccccc}
\toprule
\mrr{2}{*}{L}& \mrr{2}{*}{D?}    &         \mc{2}{c}{\bf \uftpn}                   &  \mc{2}{c}{\bf \uftsa}                        \\
              &                                & Val Acc. ($\Delta$)      & Gap                  & Val Acc.  ($\Delta$)    & Gap                 \\
\hline                                                                                                                                      
0             &                                & 78.02 (--2.19)           & \color{red}{--9.72}  & 82.81 (+2.60)          & --4.80               \\
1             &                                & 76.86 (--3.35)           & \color{red}{--11.14} & 79.56 (--0.65)          & \color{red}{--7.43} \\
1         & \cm                        & \ul{82.01} (+1.80)   & --5.63           & \ul{83.39} (+3.18)  & --2.05          \\
2             &                                & 76.32 (--3.89)           & \color{red}{--11.58} & 79.71 (--0.50)          & \color{red}{--7.23} \\
2         & \cm                        & {\bf82.43} (+2.22)   & --4.83           & {\bf 83.86} (+3.65) & --1.90          \\
\bottomrule
\end{tabular}
}
\fi
\savespace{0.07in}
\caption{\textbf{Number of projection layers (L) during finetuning, and whether
they are discarded (D) during testing.} Numbers are from Celeb-A 20-shot.
$\Delta$ denotes changes compared to no finetuning.}
\label{tab:projection}
\end{center}
\end{minipage}
\end{small}
\end{center}
\savespacebeforesection
\savespacebeforesection
\end{table}
\savespacebeforesection
\subsection{Methods for Comparison}
\savespacebeforesection
\label{sec:baselines}

As outlined in Section~\ref{sec:method}, we consider the following representation
learning and finetuning methods:
\begin{itemize}[leftmargin=*]
\savespacebeforeitem
\item \textbf{ID} trains a network to perform the auxiliary task of face
identity classification (Celeb-A only).

\savespacebeforeitem
\item \textbf{SA}, or supervised attribute, resembles the ``Baseline'' approach
in the FSL literature~\citep{closerlook}. The network learns representations by
predicting the training attributes associated with each image.

\savespacebeforeitem
\looseness=-1
\item \textbf{U} denotes unsupervised representation learning (SimCLR). We
train separate models on the Celeb-A and Zappos datasets. For ImageNet, we
utilize the off-the-shelf model checkpoint trained on the full ImageNet-1K.
\item \textbf{UFTE/UFTA} as explained in Section \ref{sec:ft}, we evaluate UFTE and UFTA which finetune on training episodes and training attributes respectively.
\end{itemize}

In addtion, we also consider a set of classic few-shot and meta-learning methods.
These methods are directly trained on \taskname{} episodes of training
attributes.
\begin{itemize}[leftmargin=*]
\savespacebeforeitem
\item \textbf{MatchingNet}~\citep{matchingnet} is a soft version of
1-nearest-neighbor. At test time, it will retrieve the label of the closest support
example in the feature space.

\savespacebeforeitem
\looseness=-1000
\item \textbf{MAML}~\citep{maml} performs several gradient descent steps in an
episode and learns the parameter initialization. For simplicity, we used the
ANIL variant~\citep{anil} that learns the last layer in the inner loop.

\savespacebeforeitem
\item \textbf{ProtoNet}~\citep{protonet} computes an average ``prototype'' for
each class and retrieves the closest one.

\savespacebeforeitem
\item \textbf{TAFENet}~\citep{tafenet} learns a meta-network that can output
task-conditioned classifier parameters.

\savespacebeforeitem
\item \textbf{TADAM}~\citep{tadam} predicts the batch normalization parameters
by using the average features of the episode. For our task we found that
conditioning on the positive examples only works the best.
\end{itemize}

In addition to the approaches above, we also provide two oracle approaches to study the upper bound to generalization to novel attributes.

\begin{itemize}[leftmargin=*]
\savespacebeforesection
\item \textbf{Oracle SA*} learns its representations by predicting all binary
attributes including both training and test ones.

\savespacebeforeitem
\item \textbf{Oracle GT} directly uses the ground-truth binary attribute values
as input features, and the readout is performed by training a logistic
regression. It still needs to select the active attributes that are used in
each episode.
\end{itemize}

\savespacebeforeitem
For the few-shot learning stage, as explained in Section \ref{sec:fsl}, we mainly use logistic regression (LR)
in few-shot episodes, but we also report results using the nearest neighbor
(NN) and nearest centroid (NC) classifiers. Note that the few-shot classifiers can be composed with any of the above representation learning methods (e.g. SA, U, UFTE, UFTA, etc.).

\savespacebeforesection
\paragraph{Implementation details:}
For Celeb-A and Zappos, images were cropped and resized to 84$\times$84. We
used ResNet-12~\citep{resnet,tadam} as the CNN backbone. The projection MLPs
have 512-512-128 units. We train SimCLR entirely on Celeb-A/Zappos images, \ie
\emph{not} using pre-trained ImageNet checkpoints for fair comparison. For
ImageNet-with-Attributes, we utilize the off-the-self SimCLR model from
ImageNet-1k, which has access to more unlabeled images. The image dimensions
are 224$\times$224. Additional details are in the Appendix.

\begin{table}[t]
\savespacefigtop{-0.5in}
\centering
\begin{minipage}[t]{0.34\textwidth}
\resizebox{\columnwidth}{!}{
\begin{tabular}{lll|cc}
\toprule
                                &         &          & \mc{2}{c}{\bf ImageNet-with-Attributes}                                \\
                                & {\it X} & {\it A}  & 5-shot                            & 20-shot                            \\
\hline 
\mc{3}{l|}{Chance}                                   & 50.00 {\sr$\pm$ 0.00}             & 50.00 {\sr$\pm$ 0.00}              \\
\hline
MAML                            &         &          & 57.90 {\sr$\pm$ 0.75}             & 57.46 {\sr$\pm$ 0.70}              \\
\mc{1}{l}{U}                    & \cm     &          & 69.05 {\sr$\pm$ 0.65}             & 71.25 {\sr$\pm$ 0.62}              \\
SA                              &         & \cm      & 64.36 {\sr$\pm$ 0.68}             & 64.16 {\sr$\pm$ 0.65}              \\
\mc{1}{l}{ \uftpn{}} &  \cm &      &  \ul{70.92} {\sr$\pm$ 0.69}    &  \ul{72.12} {\sr$\pm$ 0.63}     \\
 \uftsa{}             &  \cm &  \cm  & \textbf{71.12} {\sr$\pm$ 0.65} &  \textbf{72.91}{\sr$\pm$ 0.63}  \\
\bottomrule
\end{tabular}
}
\caption{\textbf{5- and 20-shot results on ImageNet.}
Learners uses logistic regression (LR) at test time.}
\label{tab:imagenet-main}
\end{minipage}
\hfill
\begin{minipage}[t]{.33\textwidth}
\centering
\resizebox{\columnwidth}{!}{
\begin{tabular}{l|ccc}
\toprule
             &                                                                                                              \\
             & NN                                 & NC                                 & LR                                 \\
\hline                                                                                                                 
Meta         & 61.28 {\sr$\pm$ 0.62}              & 61.50 {\sr$\pm$ 0.70}              & 57.46 {\sr$\pm$ 0.70}              \\
U            & 69.63 {\sr$\pm$ 0.59}              & 71.12 {\sr$\pm$ 0.66}              & 71.25 {\sr$\pm$ 0.62}              \\
SA           & 62.42 {\sr$\pm$ 0.62}              & 62.84 {\sr$\pm$ 0.68}              & 64.16 {\sr$\pm$ 0.65}              \\
 \uftpn{} &  \ul{69.77} {\sr$\pm$ 0.57}     &  \textbf{72.94} {\sr$\pm$ 0.61} &  \ul{72.12} {\sr$\pm$ 0.63}     \\
 \uftsa{} &  \textbf{71.55} {\sr$\pm$ 0.61} & \ul{72.42} {\sr$\pm$ 0.63}     &  \textbf{72.91} {\sr$\pm$ 0.63} \\
\hline
SA*          & 68.36 {\sr$\pm$ 0.60}              & 70.48 {\sr$\pm$ 0.66}              & 70.92 {\sr$\pm$ 0.64}              \\

\bottomrule
\end{tabular}
}
\caption{\textbf{20-shot \taskname{} on ImageNet} with different few-shot
learners.}
\label{subtab:imagenet-fsl}
\end{minipage}%
\hfill
\begin{minipage}[t]{.31\textwidth}
\centering
\resizebox{\columnwidth}{!}{
\begin{tabular}{l|ccc}
\toprule
             & Train                              & Test                               & Gap                 \\
             & attr                               & attr                               &                     \\
\hline                                                                                           

MAML         & 68.16 {\sr$\pm$ 0.59}              & 57.46 {\sr$\pm$ 0.70}              & \color{red}{-10.70} \\
U            & 76.36 {\sr$\pm$ 0.60}              & 71.25 {\sr$\pm$ 0.62}              & -5.11               \\
SA           & 69.03 {\sr$\pm$ 0.66}              & 64.16 {\sr$\pm$ 0.65}              & -4.87               \\
 \uftpn{} &  \textbf{78.31} {\sr$\pm$ 0.56} &  \ul{72.12} {\sr$\pm$ 0.63}     &  -6.19           \\
 \uftsa{} &  \ul{77.08} {\sr$\pm$ 0.62}     &  \textbf{72.91} {\sr$\pm$ 0.63} &  -4.17           \\
\hline
SA*          & 68.72 {\sr$\pm$ 0.64}              & 70.92 {\sr$\pm$ 0.64}              & 2.20                \\

\bottomrule
\end{tabular}
}
\caption{\textbf{Training vs. test attributes} of 20-shot \taskname{} on
ImageNet.}

\label{subtab:imagenet-traintest-gap}
\end{minipage}
\savespacebeforesection
\end{table}

\begin{figure*}[t]
\centering
\ifarxiv
\includegraphics[width=\textwidth,trim={0 8cm 0
0.3cm},clip]{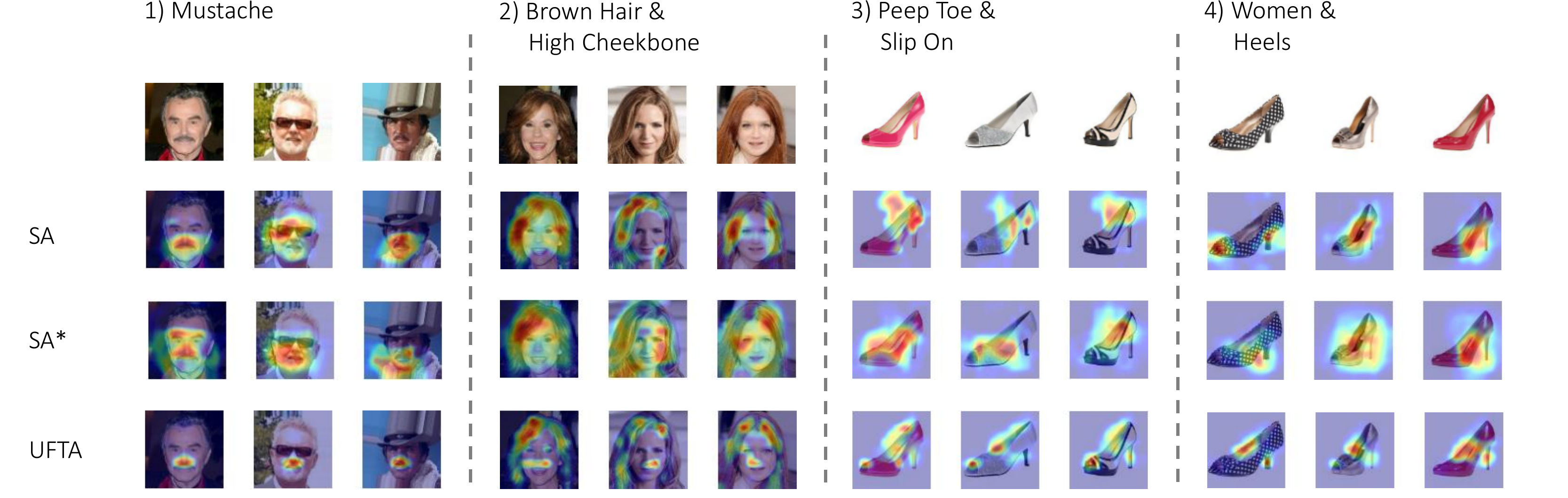}
\fi
\savespacebeforesection
\savespacebeforesection
\caption{
\textbf{Visualization of few-shot classifiers using CAM~\citep{cam}, on top of
different representations.} Left: Celeb-A; Right: Zappos-50K. Target attributes
that define the episode are shown above and images are from the query set of
the positive class at test time.}
\label{fig:viz}
\savespacebeforesection
\savespacebeforesection
\end{figure*}

\begin{figure*}[t]
\savespacefigtop{-0.5in}
\centering
\begin{minipage}[b]{0.49\textwidth}
\centering
\ifarxiv
\includegraphics[height=3.1cm,trim={0.2cm 0.2cm 0.2cm 0.2cm},clip]{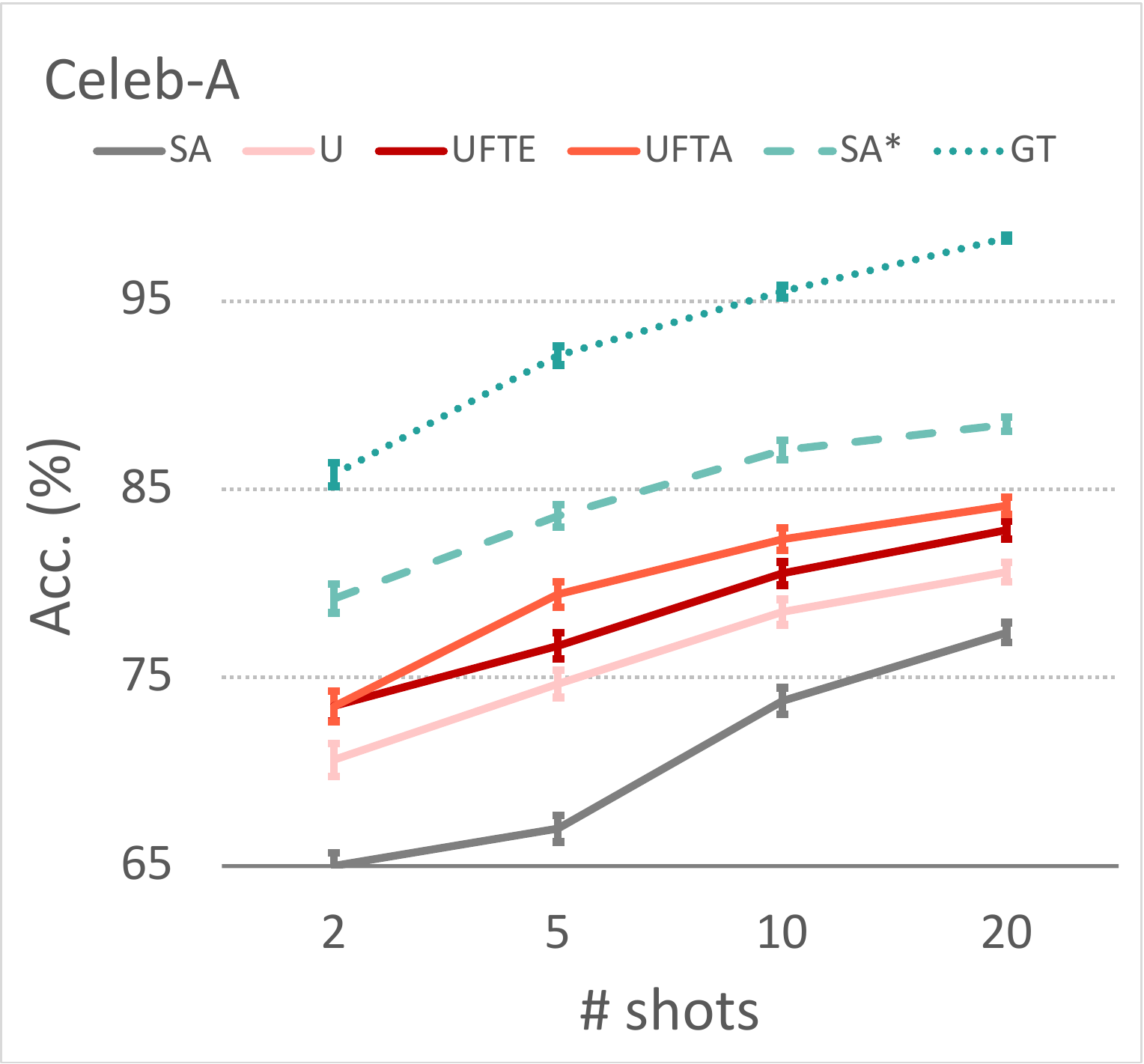}
\quad
\includegraphics[height=3.1cm,trim={0.2cm 0.2cm 0.2cm 0.2cm},clip]{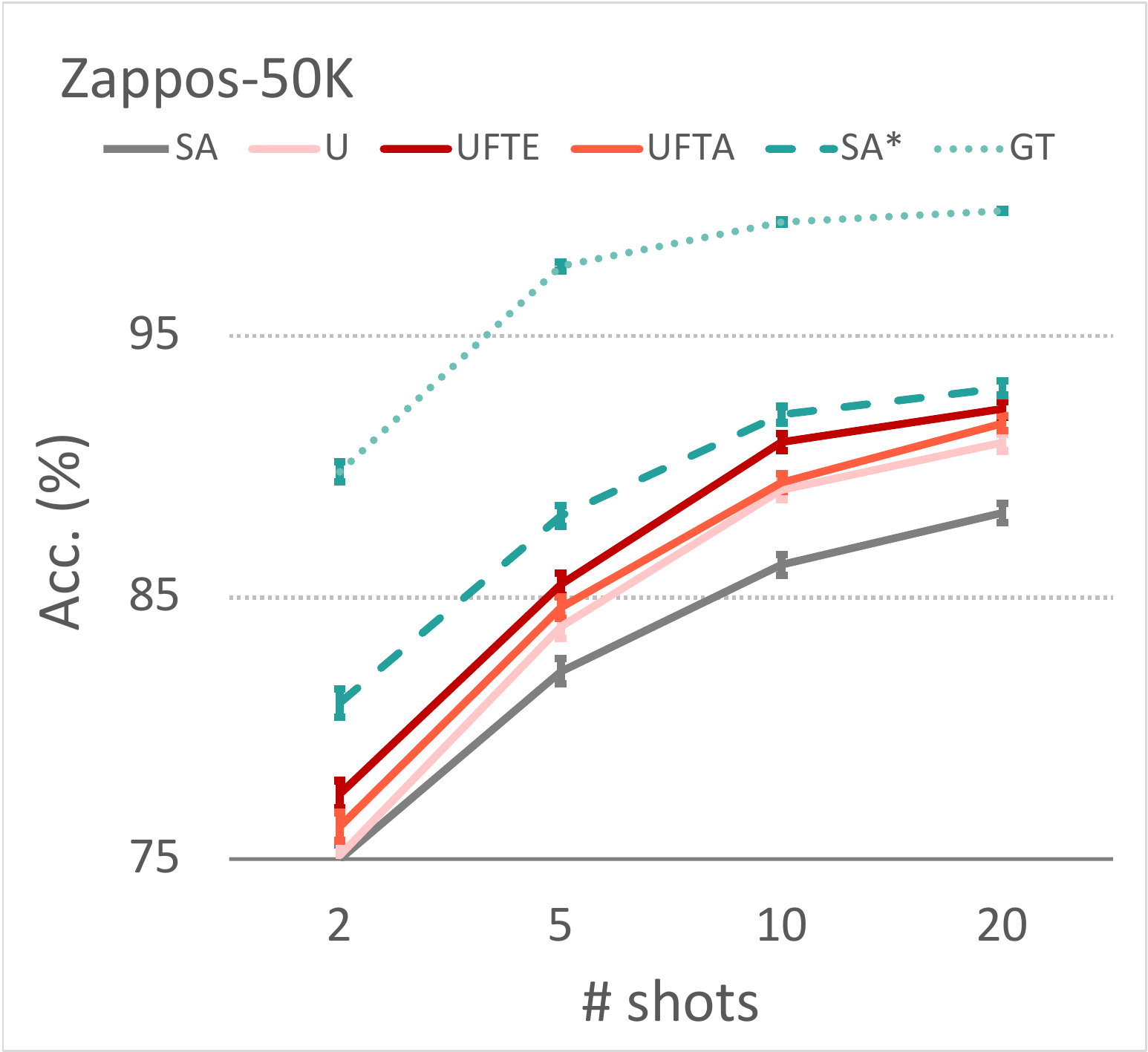}
\else
\includegraphics[height=2.8cm,trim={0.2cm 0.2cm 0.2cm 0.2cm},clip]{figures/celeb-a-nshot-v3.pdf}
\quad
\includegraphics[height=2.8cm,trim={0.2cm 0.2cm 0.2cm 0.2cm},clip]{figures/zappos-nshot-v4.pdf}
\fi
\savespacebeforesection
\caption{\textbf{How many examples are needed for \taskname{}?} Performance
increases with number of shots, even when given the binary ground-truth
attribute vector (\textbf{GT}), suggesting that there is greater ambiguity in \taskname{} than in standard FSL.}
\label{fig:nshot}
\end{minipage}
\hfill
\begin{minipage}[b]{0.49\textwidth}
\centering
\ifarxiv
\includegraphics[height=3.3cm]{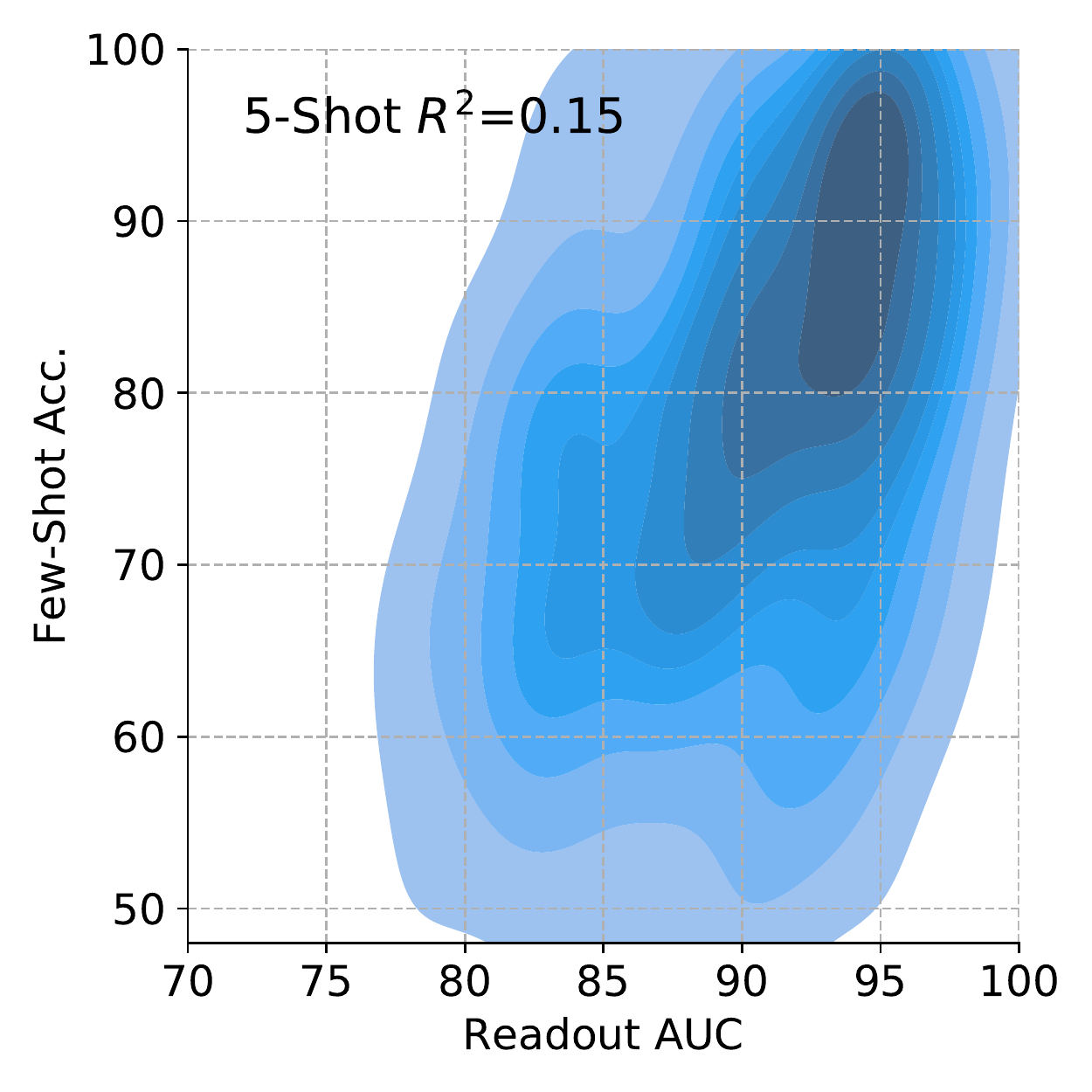}
\quad
\includegraphics[height=3.3cm]{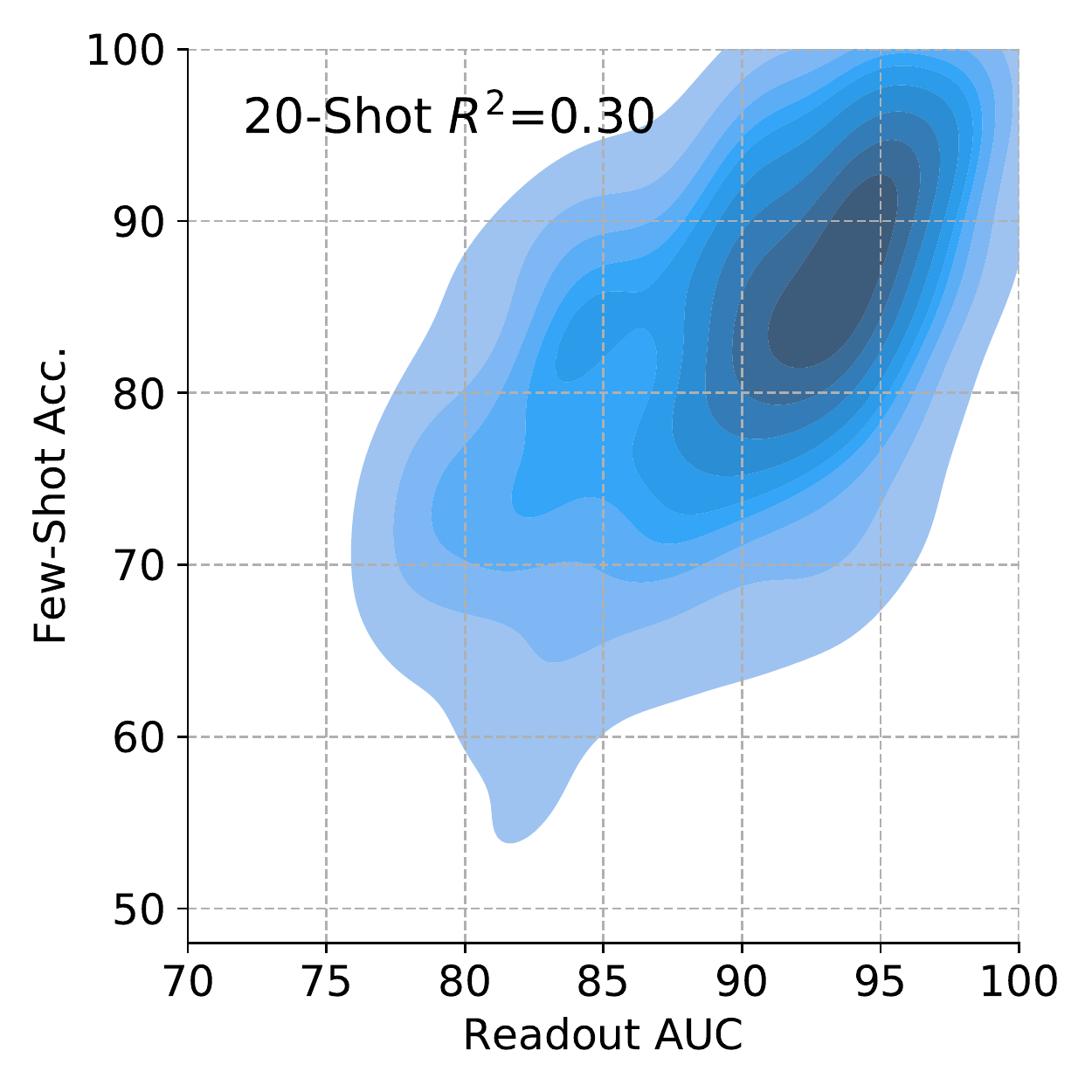}
\else
\includegraphics[height=3.0cm]{figures/test-s5.pdf}
\quad
\includegraphics[height=3.0cm]{figures/test-s20.pdf}
\fi
\vspace{-0.08in}\savespacebeforesection
\caption{\textbf{Correlation between readout AUC and few-shot acc. using \uftsa}. Variance can be explained by the challenge of predicting attributes and the ambiguity of \taskname{}. More shots reduce variance and
improve correlation.}
\label{fig:corr}
\end{minipage}
\vspace{-0.15in}
\end{figure*}

\savespacebeforesection
\subsection{Comparing Various Methods for \taskname{}}
\savespacebeforesection
\label{sec:experiments:results}

Table~\ref{tab:main} shows our main results on Celeb-A and Zappos-50K with 5-
and 20-shot episodes. Table~\ref{tab:combo} explores different combinations of
representations and few-shot learners. Overall, the standard episodic
meta-learners performed relatively poorly. Also, supervised attribute (SA)
learning and learning via the auxiliary task of class facial identification
(ID) were not helpful for representation learning either. Interestingly, U
attained relatively better test performance, suggesting that the training
objective in contrastive learning indeed preserves more general features---not
just for semantic classification tasks as shown in prior work, but also for the
flexibly-defined attribute classes in our FSAL paradigm. All these evidences suggest that unsupervised representation learning is better than supervised methods for FSAL. 

Moreover, \uftsa{} and \uftpn{} approaches obtained significant gains in
performance, suggesting that a combination of unsupervised features with some
supervised information is indeed beneficial for this task. Lastly, they
are able to reduce the generalization gap between SA and the oracle SA*, in
fact almost closing it entirely on Zappos-50K. We investigate and analyze the benefit of unsupervised pretraining and supervised finetuning further in Section~\ref{sec:analysis}.

Results on ImageNet-with-Attributes are reported separately for clarity,
because U, \uftpn{}, and \uftsa{} had access to additional unlabeled examples.
As shown in Table~\ref{tab:imagenet-main}, both \uftpn{} and \uftsa{}
outperformed other methods substantially. Because of the additional unlabeled
data available in this setting, even U achieved a substantially better accuracy
than SA and MAML. Results in Table~\ref{subtab:imagenet-fsl} show that \uftpn{}
and \uftsa{} work well when combined with different few-shot learners.

\savespacebeforesection
\looseness=-10000
\paragraph{Visualizing few-shot classifiers:} To understand and interpret the
decision made by few-shot linear classifiers, we visualize the classifier
weights by using CAM~\citep{cam}, and plot the heatmap over the 11$\times$11
spatial feature map in Figure~\ref{fig:viz}. SA sometimes shows incorrect
localization as it is not trained to classify those novel test attributes. SA*
shows bigger but less precise heatmaps since the training objective encourages
the propagation of attribute information spatially. In contrast, UFTA produces
accurate and localized heatmaps that pinpoint the location of the attributes
(e.g. mustache or cheekbone); this is impressive since no labeled information
concerning these attributes was available during representation pre-training
and finetuning. This result supports the hypothesis that local features can be
good descriptors that match different views of the same instance during
contrastive learning, and finetuning further establishes a positive transfer
between training and test attributes.

\savespacebeforesection
\paragraph{Number of shots and task ambiguity:} 
Our few-shot attribute learning episodes can be ambiguous. For example, by
presenting only a smiling face with eyeglasses in the support set, it is
unclear whether the positive set is determined by ``smiling'' or ``wearing
eyeglasses''. Figure~\ref{fig:nshot} show several approaches evaluated using LR
with varying numbers of support examples per class in Celeb-A and Zappos-50K
episodes, respectively. The oracle GT gradually approached 100\% accuracy as
the number of shots approached 20. This demonstrates that \taskname{} tasks
potentially require more support examples than standard FSL to resolve
ambiguity. Again here, \uftsa{} and \uftpn{} consistently outperformed U, SA,
and ID across different number of shots. Figure~\ref{fig:corr} shows the
correlation between readout performance of attributes and few-shot learning
accuracy, using \uftsa{}. With a larger number of shots, there is a higher
correlation between the two, but there is still a large amount of variance that
is due to the ambiguity of the task itself. More details are included in the
Appendix.

\savespacebeforesection
\paragraph{Ablation studies:} Table~\ref{tab:projection} studies the effect of
the projection MLP for attribute classification finetuning. Adding MLP
projection layers was found to be beneficial for unsupervised learning in prior
work~\citep{simclr}. Here we found that adding MLP layers is also critical in
supervised finetuning. Finetuning directly on the
backbone (depth=0), and keeping the MLP during test (Discard=no) both led to
significant drop in performance. In the Appendix, we also report on studies of
the effect of adding the L1 regularizer on LR.

\savespacebeforesection
\subsection{Analysis of Few-Shot Generalization}
\label{sec:analysis}
\savespacebeforesection
In Tables~\ref{tab:gap} and \ref{subtab:imagenet-traintest-gap}, we study the
performance gap between training attributes and test attributes. Notably, SA
performs very well on test episodes defined using training attributes, but
there is a large generalization gap between training and test attributes.
\uftpn{} and \uftsa{} show significant improvements in terms of reducing the
generalization gap between training and test attributes. Moreover, we find that
self-supervised pre-training generally preserves informative features and is
more general than supervised pre-training.

\savespacebeforesection
\paragraph{Investigating the cause of generalization issues:} 
\looseness=-1
We hypothesize that the weak performance of episodic learners and SA on our
benchmarks is because their training objectives essentially encourage ignoring
attributes that are not useful at training time, but may still be useful at
test time. In Appendix~\ref{app:toy_problem}, we study a synthetic problem to
further analyze these generalization issues. We explore training a ProtoNet
model on data from a linear generative model, where each FSAL episode presents
ambiguity in identifying the relevant attributes. In this setting, the network
is forced to discard information that is useful for test tasks to solve the
training tasks effectively, and thus fails to generalize.

\begin{figure*}[t]
\savespacefigtop{-0.5in}
\centering
\includegraphics[width=\textwidth]{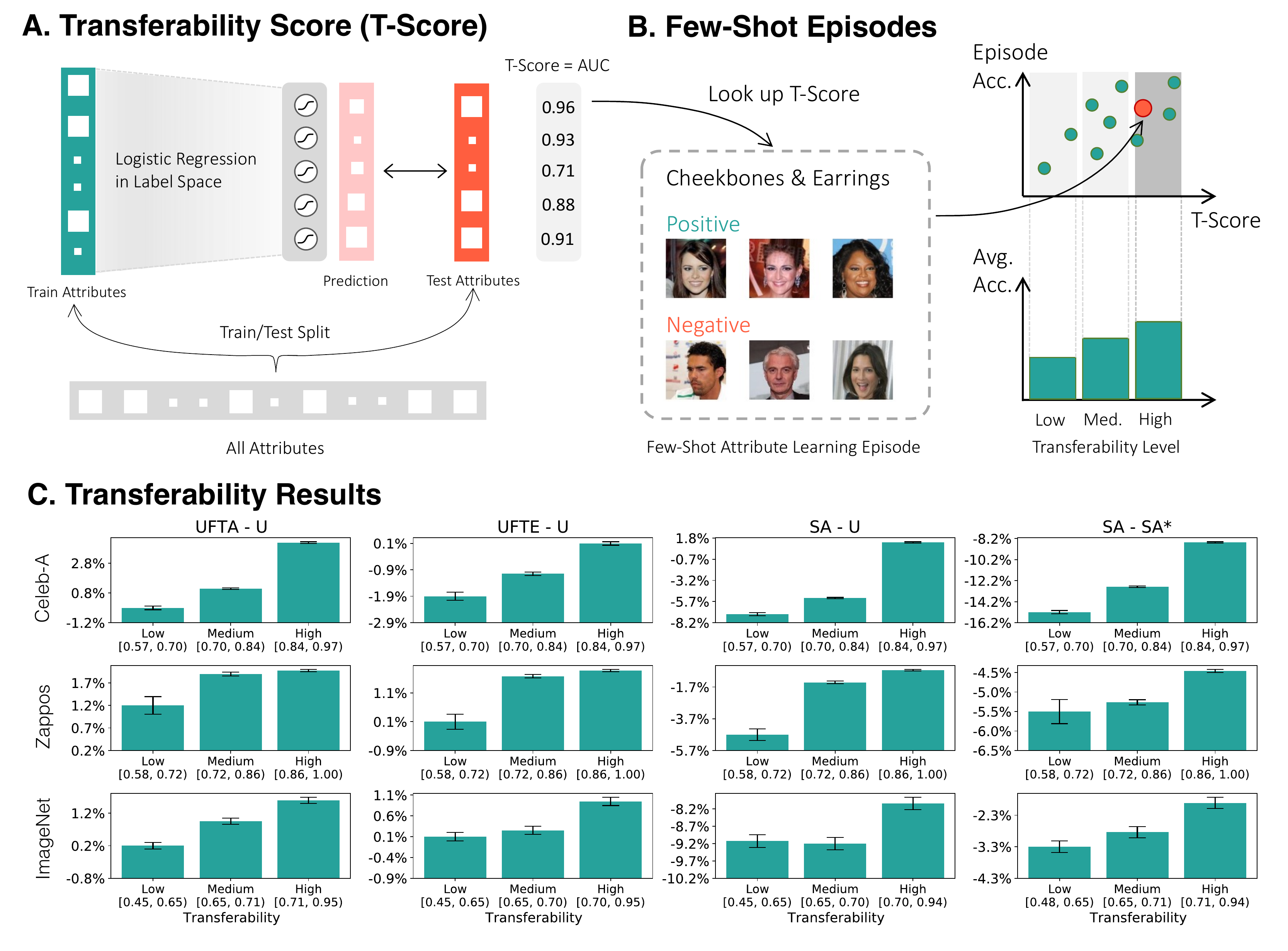}
\savespacebeforesection
\savespacebeforesection
\savespacebeforesection
\caption{
\looseness=-10000
\textbf{Few-shot performance vs. transferability across training and test
attributes.} \textbf{A:} Transferability score (T-score) is computed based on
the AUC of a test attribute predicted by a logistic regression model on a set
of training attributes. 100 different random splits across train/test
attributes per split are used. \textbf{B:} Both episodic accuracy and T-scores
are recorded on 60,000 episodes (600 episodes per split). Episodes are grouped
into three bins by their T-scores. \textbf{C:} Performance of training or
finetuning on training attributes correlates with T-score. Error bars are
standard errors in each bin.}
\label{fig:transfer}
\savespacebeforesection
\savespacebeforesection
\end{figure*}

\savespacebeforesection
\paragraph{Transferability score:}
\looseness=-1000
We aim to investigate the question of why unsupervised pretraining and
supervised finetuning produce better performance, and whether the performance
difference is caused by the closeness between training and test attributes.
More concretely, we aim to predict the transferability between training and
test splits by analyzing the training vs. test attributes. Each image has a
complete attribute vector, describing the values of each attribute in the
image. Some of these attributes are in the training set, and others in the test
set. To quantify the transferability, we leverage the idea of mutual
information. In particular, we learn a logistic regression model that takes the
training attribute vector in a particular image as input and predicts the value
of one of the test attributes in that image. Each logistic regression model
will generate an AUC score on held-out images, and we average them across the
relevant test attributes in each episode, and we define this AUC score as the
``transferability score.'' Our hypothesis is that more mutual information
between the attribute label distributions will translate to higher transfer
performance.

In Figure~\ref{fig:transfer}, we ran experiments using 100 random splits of
training and test attributes. The results verify our hypothesis. We see
positive correlation between the transfer performance and our transferability
score: When subtracting U as a baseline, both UFTA and UFTE models get better
when there is higher transferability (subtraction reduces the effect of
per-episode variability). The same conclusion can be drawn when we subtract SA*
from SA. By plotting the relation between U and SA, we show that supervised
learning is more helpful when there is higher transferability in the label
space whereas self-supervised learning is more flexible at adapting to novel
target tasks.

\looseness=-10000
To summarize, our empirical evidence suggests that unsupervised representation
learning is superior to supervised methods in terms of retaining information
relevant to the test attributes. Other methods, such as supervised
representation learning or episodic training, tend to effectively ignore
attributes that were not used for labelling during training. Moreover,
supervised finetuning of the representations is helpful when the
transferability between test and train attributes is high, but less so
otherwise, and supervised pretraining alone is harmful to novel attribute
generalization for most train vs. test attribute pairs.
\savespacebeforesection
\section{Conclusion}
\savespacebeforesection
\savespacebeforeitem
\looseness=-1

To investigate few-shot generalization, we developed FSAL, a novel few-shot
learning paradigm that requires learners to generalize to novel attributes at
test time. We developed benchmarks using the Celeb-A, Zappos-50K, and ImageNet
datasets to create learning episodes using existing attribute labels. This
setting presents a strong generalization challenge, since the split in
attribute space can make the training and test tasks less similar than
traditional few-shot learning tasks. Consequently, standard supervised
representation learning performs poorly on the test set, unlike recent
benchmark results in few-shot learning of semantic classes. However,
unsupervised contrastive learning preserved more general features, and further
finetuning yielded strong performance. We also studied the performance gap
under different splits in the attribute label space where we found that
supervised representation learning works better when there is more information
shared between train and test attributes.

\textbf{Limitations:} Our empirical analysis could be made more complete by
including other unsupervised representation learning methods and extending to
other domains. Further, the episodes contained in our benchmark tasks can
sometimes be difficult for humans to resolve even after we removed ambiguous
attributes.

\textbf{Societal Impact:} FSAL relies on attribute labels, which can be
difficult to obtain and encode bias in some settings (\eg the \emph{attractive}
attribute in Celeb-A).

\ifarxiv
\section*{Contribution Statement}
All authors contributed to the high-level idea and writing of the paper. MR contributed to the code base for running attribute-based few-shot learning experiments, discovered that unsupervised learning plus finetuning is beneficial, performed experiments on Celeb-A, and created most of the figures and graphics. ET helped with figure creation, implemented the flexible few-shot version of Zappos-50K and ran the experiments on that dataset. KCW contributed to the FFSL task definition, implemented the ImageNet-with-Attributes FFSL benchmark, and the associated code for using the off-the-shelf models. JL designed and implemented early experiments in the FFSL setting, provided the formal description of FFSL, and analyzed the linear toy problem presented in Appendix E. JS contributed to the analysis of early FFSL experiments. XP and AT contributed ideas about the underlying question and its possible solutions, and helped interpret results. RZ contributed many ideas behind the underlying question studied here and the
problem formulation, and led the team's brainstorming about how to test the hypotheses, the datasets and benchmarks, and modeling approaches and visualizations.

\section*{Acknowledgments}
We would like to thank Claudio Michaelis for several helpful discussions about the problem formulation, Alireza Makhzani for related generative modeling ideas, and Mike Mozer for discussions about contextual similarity. Resources used in preparing this research were provided, in part, by the Province of Ontario, the
Government of Canada through CIFAR, and companies sponsoring the Vector Institute
(\url{www.vectorinstitute.ai/\#partners}). This project is supported by NSERC and the Intelligence
Advanced Research Projects Activity (IARPA) via Department of Interior/Interior Business Center
(DoI/IBC) contract number D16PC00003. The U.S. Government is authorized to reproduce and distribute
reprints for Governmental purposes notwithstanding any copyright annotation thereon. Disclaimer: The
views and conclusions contained herein are those of the authors and should not be interpreted as
necessarily representing the official policies or endorsements, either expressed or implied, of
IARPA, DoI/IBC, or the U.S. Government.
\fi
\bibliography{ref}
\appendix
\clearpage
\appendix
\ifarxiv
\clearpage
\fi
\section{Attribute Readout}
In Tab.~\ref{tab:attrreadout} and \ref{tab:attrreadout-imageneta}, we provide
attribute readout performance with different learned representations. This is a
similar task that measures the generalizability, but it does not evaluate the
rapid learning aspect brought by few-shot learning.

\begin{table}[t]
\begin{small}
\begin{center}
\begin{tabular}{lccccccc|c}
\toprule
{\bf Mean AUC}  & {\bf RND} & {\bf PN}  & {\bf ID}  & {\bf SA}  & {\bf U}   & \gr {\bf \uftpn} & \gr {\bf \uftsa}   & {\bf SA*}  \\
\hline      
All (40)        &  79.18    &  88.80      & 91.29     & 90.27     & 92.80   & \gr {\bf 93.34}  & \gr {\bf 93.33}      & 94.46      \\
Train+Test (27) &  82.27    &  93.38      & 94.31     & 94.23     & 95.78   & \gr {\bf 96.53}  & \gr {\bf 96.52}      & 97.18      \\
Train (14)      &  84.40    &  96.04      & 95.34     & 96.04     & 96.43   & \gr {\bf 97.23}  & \gr {\bf 97.23}      & 97.50      \\
Test  (13)      &  79.96    &  90.52      & 93.19     & 92.63     & 95.08   & \gr {\bf 95.78}  & \gr {\bf 95.76}      & 96.84      \\
\bottomrule
\end{tabular}
\end{center}
\end{small}
\caption{\textbf{Celeb-A attribute readout} performance of different
representations, measured in mean AUC. RND denotes using a randomly initialized
CNN; PN denotes ProtoNet.}
\label{tab:attrreadout}
\end{table}

\section{Ablation studies}
Table~\ref{tab:l1} studies the effect of the L1 regularization. The benefit is
especially noticeable on SA* and GT, since it allows the few-shot learner to
have a sparse selection of disentangled feature dimensions.

\begin{table}
\begin{small}
\begin{center}
\resizebox{0.8\columnwidth}{!}{
\begin{tabular}{cllll|ll}
\toprule
                & {\bf SA}              & {\bf U}                & {\bf \uftpn}         & {\bf \uftsa}           & {\bf SA*}            & {\bf GT}            \\
\hline  
LR              & 77.4                  & 79.2                   & 82.2                 & 83.1                   & 87.1                 & 95.8                \\
\gr +L1 (1e-4)  & \gr 77.6 (+0.2)       & \gr 79.4 (+1.2)        & \gr 82.3 (+0.1)      & \gr 83.2 (+0.1)        & \gr 87.4 (+0.3)      & \gr 96.1 (+0.3)     \\
\gr +L1 (1e-3)  & \gr {\bf78.2} (+0.8)  & \gr {\bf80.2} (+1.0)   & \gr {\bf 82.4} (+0.2)& \gr {\bf83.8} (+0.7)   & \gr {\bf88.4} (+1.3) & \gr 97.1 (+1.3)     \\
\gr +L1 (1e-2)  & \gr 75.7 (\gt{--1.7}) & \gr 78.3 (\gt{--0.9})  & \gr 78.8 (\gt{--3.5})      & \gr 79.5 (\gt{--3.6})  & \gr 87.6 (+0.5)      & \gr {\bf98.2} (+2.4)\\
\bottomrule
\end{tabular}
}
\end{center}
\end{small}
\caption{\textbf{Effect of the L1 regularizer} on different representations for
the validation set of Celeb-A 20-shot.}
\label{tab:l1}
\end{table}

\begin{table}
\begin{small}
\begin{center}
\begin{tabular}{lcccc}
\toprule
{\bf Mean AUC}             & {\bf SA}  & {\bf SA*}  & {\bf U}   & {\bf UFTA}   \\
\hline

\hline
 All (25 attributes)        & 72.01 & 73.02 & 81.08 & \textbf{82.49} \\
 Train+Test (21 attributes) & 73.43 & 78.98 & 80.14 & \textbf{82.37} \\
 Train (11 attributes)      & 72.69 & 75.86 & 80.63 & \textbf{82.43} \\
 Test (10 attributes)       & 72.01 & 74.98 & 81.08 & \textbf{83.30} \\
\hline
\end{tabular}
\end{center}
\end{small}
\caption{ImageNet-with-Attributes attribute readout binary prediction
performance of different representations, measured in mean AUC. }
\label{tab:attrreadout-imageneta}
\end{table}

\section{Additional heatmap visualization}

\begin{figure*}[t]
\centering
\includegraphics[width=0.9\linewidth]{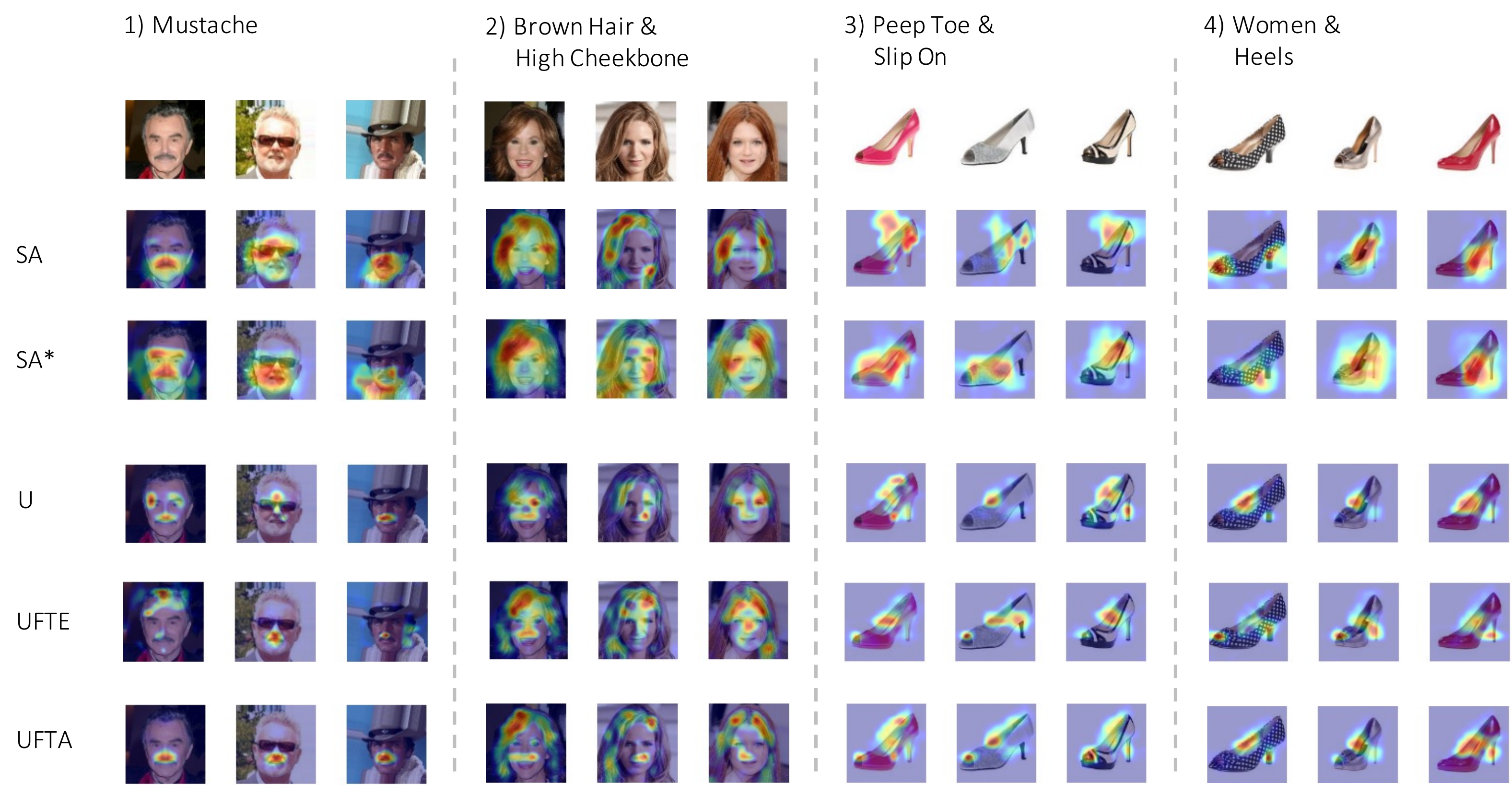}
\caption{Additional visualization results, on 20-shot episodes, including more
methods for comparison.}
\label{fig:additional_combo}
\end{figure*}

\begin{figure*}[t]
\centering
\includegraphics[width=0.9\linewidth]{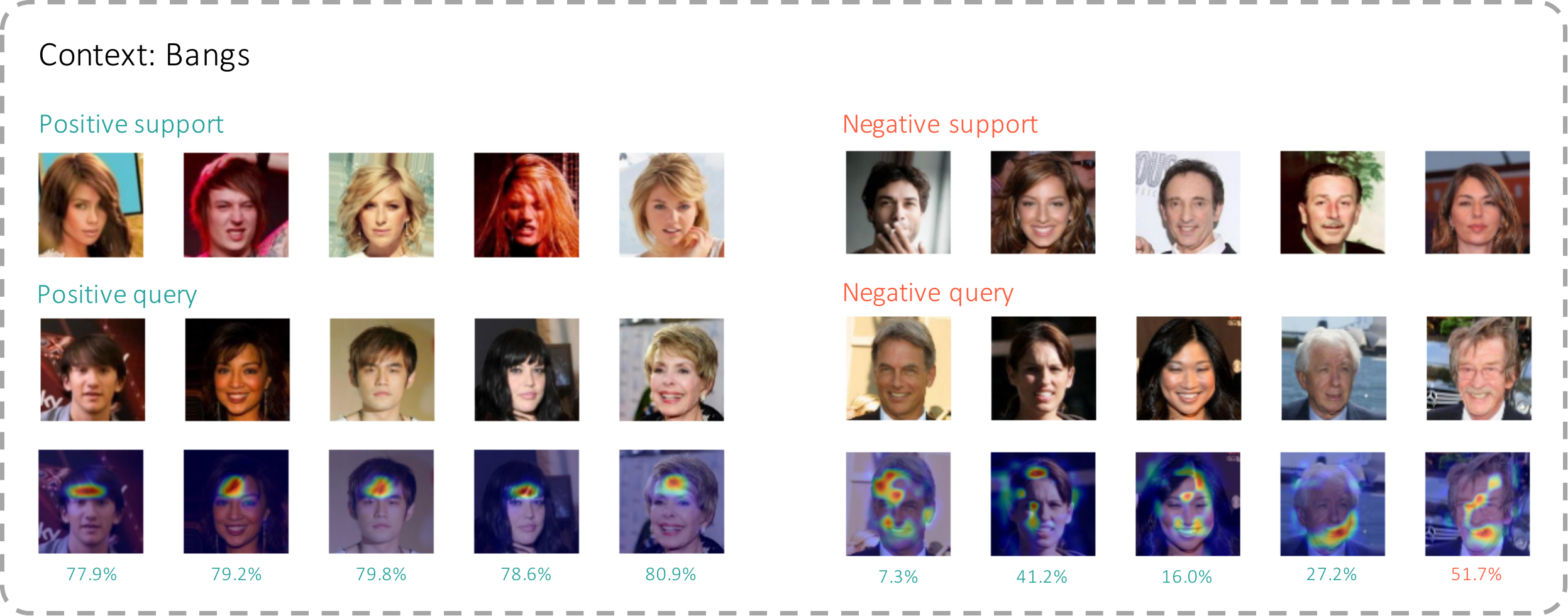}\\
\quad \\
\includegraphics[width=0.9\linewidth]{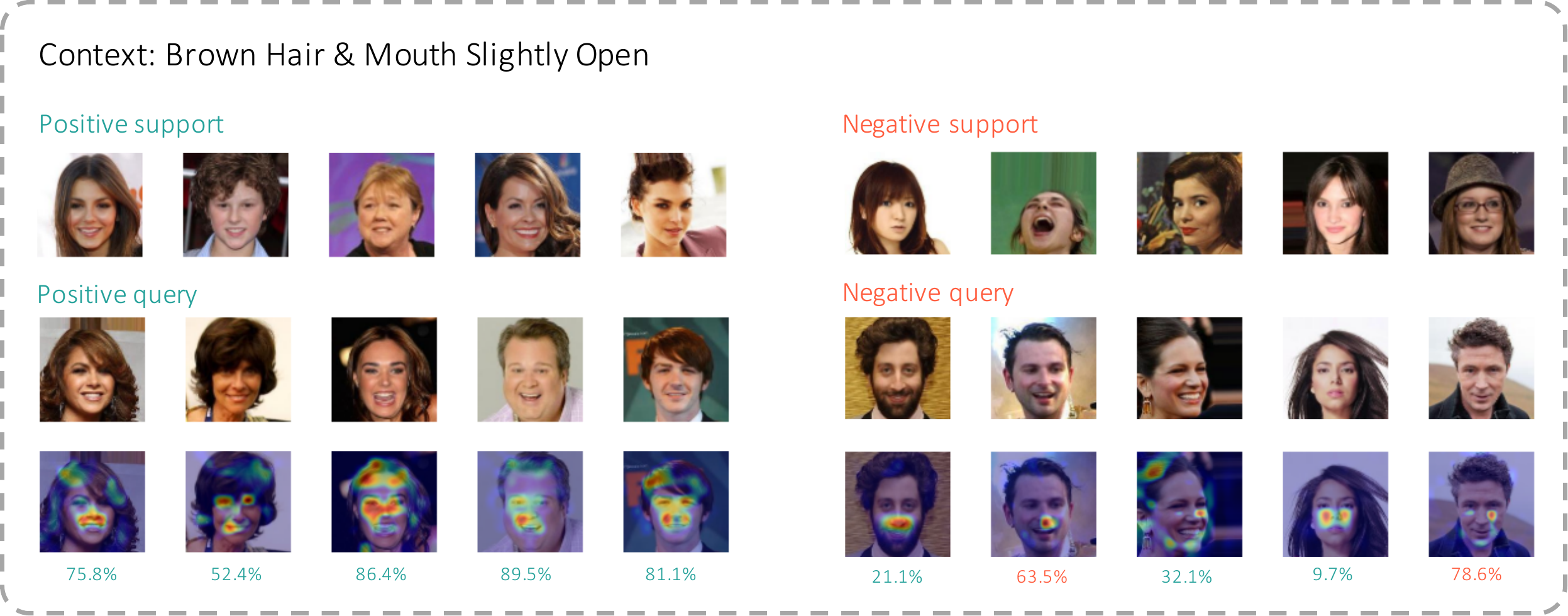}
\caption{\textbf{Visualization of Celeb-A 20-shot LR classifiers using CAM on
top of UFTA representations.} Context attributes that define the episode are
shown above. Classifier sigmoid confidence scores are shown at the bottom. Red
numbers denote wrong classification and green denote correct. }
\label{fig:celeb-a-additional}
\end{figure*}

\begin{figure*}[t]
\centering
\includegraphics[width=0.9\linewidth]{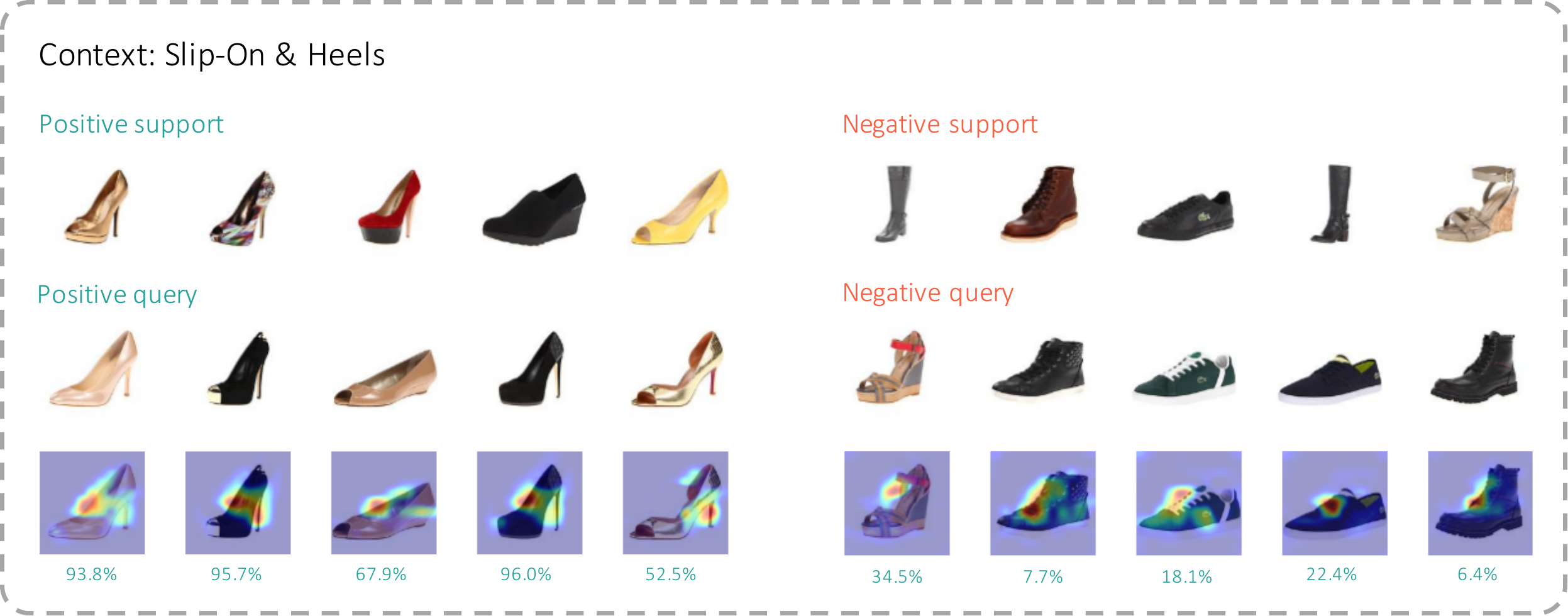}\\
\quad \\
\includegraphics[width=0.9\linewidth]{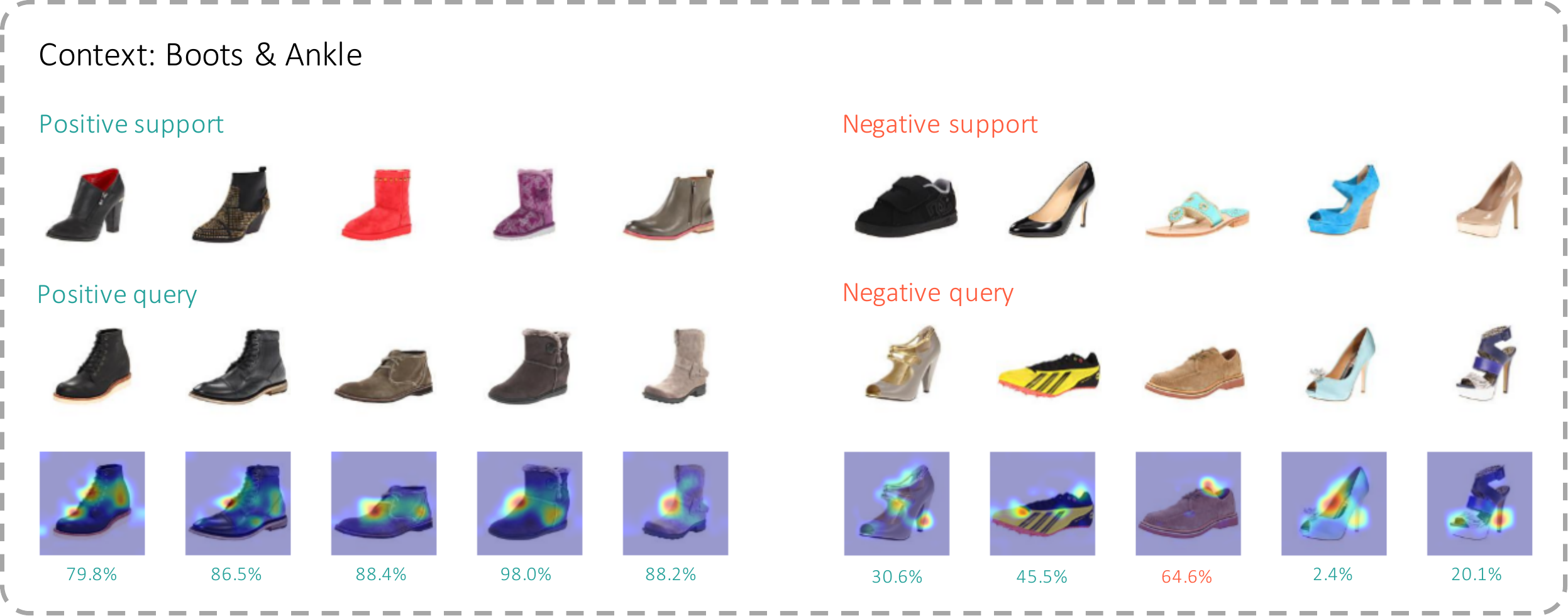}
\caption{\textbf{Visualization of Zappos-50K 20-shot LR classifiers using CAM
on top of UFTA representations.} Context attributes that define the episode are
shown above. Classifier sigmoid confidence scores are shown at the bottom. Red
numbers denote wrong classification and green denote correct. }
\label{fig:zappos-additional}
\end{figure*}

We provide additional visualization results in
Figure~\ref{fig:additional_combo}, \ref{fig:celeb-a-additional}, and
\ref{fig:zappos-additional}, and we plot the heat map to visualize the LR
classifier weights. Figure~\ref{fig:additional_combo} includes SA*, U, and UFTE
which are omitted in the main paper due to space limitations.
Figure~\ref{fig:celeb-a-additional} and \ref{fig:zappos-additional} visualize
more information including both support and query examples in the episode, and
some of the episodes are challenging to solve given just a few examples.


\begin{table*}[t]
\centering
\begin{center}
\begin{small}
\begin{tabular}{ccccc}
\hline
\mr{4}{\textbf{Train}} &
5\_o\_Clock\_Shadow & 
Black\_Hair & 
Blond\_Hair & 
Chubby\\
& 
Double\_Chin & 
Eyeglasses & 
Goatee & 
Gray\_Hair\\
& 
Male & 
No\_Beard & 
Pale\_Skin & 
Receding\_Hairline\\
& 
Rosy\_Cheeks & 
Smiling &  &  \\
\hline
\mr{4}{\textbf{Val/Test}} &
Bald & 
Bangs & 
Brown\_Hair & 
Heavy\_Makeup\\
& 
High\_Cheekbones & 
Mouth\_Slightly\_Open & 
Mustache & 
Narrow\_Eyes\\
& 
Sideburns & 
Wearing\_Earrings & 
Wearing\_Hat & 
Wearing\_Lipstick\\
& 
Wearing\_Necktie & &  &  \\
\hline
\end{tabular}
\end{small}
\end{center}
\caption{Attribute Splits for Celeb-A}
\label{tab:celebasplit}
\end{table*}

\section{Attribute splits of Celeb-A}
\label{app:split}
We include the attribute split for Celeb-A in Table~\ref{tab:celebasplit}. There are 14 attributes
in training and 13 attributes in val/test. We discarded the rest of the 13 attributes in the
original datasets since they are either hard to classify with the oracle classifier (e.g. big lips, oval face) or simply ambiguous (e.g. young, attractive).
\begin{table*}[t]
\centering
\begin{center}
\begin{small}
\resizebox{0.98\textwidth}{!}{
\begin{tabular}{ccccc}
\hline
\mr{10}{\textbf{Train}} &
Category-Shoes &
Category-Sandals &
SubCategory-Oxfords &
SubCategory-Heel \\
&
SubCategory-Boot &
SubCategory-Slipper Flats & SubCategory-Short heel& SubCategory-Flats \\
&
SubCategory-Slipper Heels & SubCategory-Athletic &
SubCategory-Knee High &
SubCategory-Crib Shoes \\
&
SubCategory-Over the Knee &
HeelHeight-High heel &
Closure-Pull-on &
Closure-Ankle Strap \\
&
Closure-Zipper & 
Closure-Elastic Gore &
Closure-Sling Back &
Closure-Toggle \\
&
Closure-Snap &
Closure-T-Strap &
Closure-Spat Strap &
Gender-Men \\ 
&
Gender-Boys & 
Material-Rubber &
Material-Wool & 
Material-Silk \\
&
Material-Aluminum &
Material-Plastic &
Toestyle-Capped Toe &
Toestyle-Square Toe \\
&
Toestyle-Snub Toe &
Toestyle-Bicycle Toe &
Toestyle-Open Toe &
Toestyle-Pointed Toe \\
&
Toestyle-Almond &
Toestyle-Apron Toe &
Toestyle-Snip Toe &
Toestyle-Medallion\\
\hline

\mr{10}{\textbf{Val/Test}} &
Category-Boots & 
Category-Slippers & 
SubCategory-Mid-Calf & 
SubCategory-Ankle \\
&
SubCategory-Loafers & 
SubCategory-Boat Shoes & 
SubCategory-Clogs and Mules &
SubCategory-Sneakers and Athletic Shoes \\
&
SubCategory-Heels & 
SubCategory-Prewalker &
SubCategory-Prewalker Boots & SubCategory-Firstwalker \\
&
HeelHeight-Short heel &
Closure-Lace up & 
Closure-Buckle & 
Closure-Hook and Loop \\
&
Closure-Slip-On &
Closure-Ankle Wrap &
Closure-Bungee &
Closure-Adjustable \\
&
Closure-Button Loop &
Closure-Monk Strap &
Closure-Belt &
Gender-Women \\
&
Gender-Girls &
Material-Suede &
Material-Snakeskin &
Material-Corduroy \\
&
Material-Horse Hair &
Material-Stingray &
Toestyle-Round Toe &
Toestyle-Closed Toe \\
&
Toestyle-Moc Toe &
Toestyle-Wingtip &
Toestyle-Center Seam &
Toestyle-Algonquin \\
&
Toestyle-Bump Toe &
Toestyle-Wide Toe Box &
Toestyle-Peep Toe & \\
\hline
\end{tabular}
}
\end{small}
\end{center}
\caption{Attribute splits for Zappos-50K}
\label{tab:zappossplit}
\end{table*}

\section{Attribute splits of Zappos-50K}
The Zappos-50K dataset annotates images with different values relating to the following aspects of shoes: `Category', `Subcategory', `HeelHeight', `Insole', `Closure', `Gender', `Material' and `Toestyle'.

We discarded the `Insole' values, since those refer to the inside part of the shoe which isn't visible in the images. We also discarded some `Material' values that we deemed hard to recognize visually. We also modified the values of `HeelHeight' which originally was different ranges of cm of the height of the heel of each shoe. Instead, we divided those values into only two groups: `short heel' and `high heel', to avoid having to perform very fine-grained heel height recognition which we deemed was too difficult.

These modifications leave us with a total of 79 values (across all higher-level categories). Not all images are tagged with a value from each category, while some are even tagged with more than one value from the same category (e.g. two different materials used in different parts of the shoe). We split these values into 40 `training attributes' and 39 `val/test attributes'. 

We include the complete list of attributes in Table~\ref{tab:zappossplit}. The format we use is `X-Y' where X stands for the category (e.g. `Material') and Y stands for the value of that category (e.g. `Wool'). We do this to avoid ambiguity, since it may happen that different categories have some value names in common, e.g. `Short Heel' is a value of both `SubCategory' and `HeelHeight'.


\begin{table*}[t]
\centering
\begin{center}
\begin{small}
\begin{tabular}{ccccc}
\hline
\mr{4}{\textbf{Train}} &
pink & 
spotted & 
wet & 
blue\\
& 
shiny & 
rough & 
striped & 
white\\
& 
metallic & 
wooden & 
gray & 
\\
\hline
\mr{4}{\textbf{Val/Test}} &
brown & 
green & 
violet & 
red\\
& 
orange & 
yellow & 
furry & 
black\\
& 
vegetation & 
smooth & 
 & 
\\
\hline
\end{tabular}
\end{small}
\caption{Attribute Splits for ImageNet-with-Attributes}
\label{tab:imageneta-split}
\end{center}
\end{table*}

\section{Attribute splits of ImageNet-with-Attributes}
\label{app:imageneta-split}
We include the attribute split for ImageNet-with-Attributes in Table~\ref{tab:imageneta-split}.
There are 11 attributes in training and 10 attributes in val/test. We discarded the rest of the 4
attributes in the ``shape'' category (long, round, rectangular and square), since they are difficult to predict from the images.
\section{Few-Shot Attribute Learning Toy Problem}
\label{app:toy_problem}

In this section, we present a toy problem that illustrates the challenges introduced by the FSAL setting and the failures of existing approaches on this task. This simple problem
captures the core elements of our FSAL tasks, including ambiguity, introducing novel attributes at test time, and the role of learning good representations. The primary limitation of this
model is the fact that it is fully linear and the attribute values are independent---in a more
realistic FSAL task recovering a good representation from the data is significantly more
challenging, and the data points will have a more complex relationship with the attributes as in our
benchmark datasets.

\paragraph{Problem setup} We define a FSAL problem where the data points $\bx
\in \bbR^{m}$ are generated from binary attribute strings, $\bz \in \{0, 1\}^d$, with $\bx = A \bz +
\bzeta$ for some matrix $A \in \bbR^{m \times d}$ with full column rank and noise source $\bzeta$.
Thus, each data point $\bx$ is a sum of columns of $A$ with some additive noise.

In each episode, examples are labelled as positive when two designated entries of the attribute strings are both 1-valued,
and negative otherwise. For the training episodes, the labels depend only on the first $d_1 < d$
entries of $\bz$. At test time, the labels depend on the remaining $d -d_1$ attributes. The training and test
episodes are generated by choosing two of the attributes in the respective sets. Then $k$ data points are sampled with positive labels (the two attributes are 1-valued) and $k$ with negative labels (at least one of the attributes is
0-valued).

\begin{figure}
    \begin{minipage}{0.48\linewidth}
    \centering
    \includegraphics[width=\linewidth]{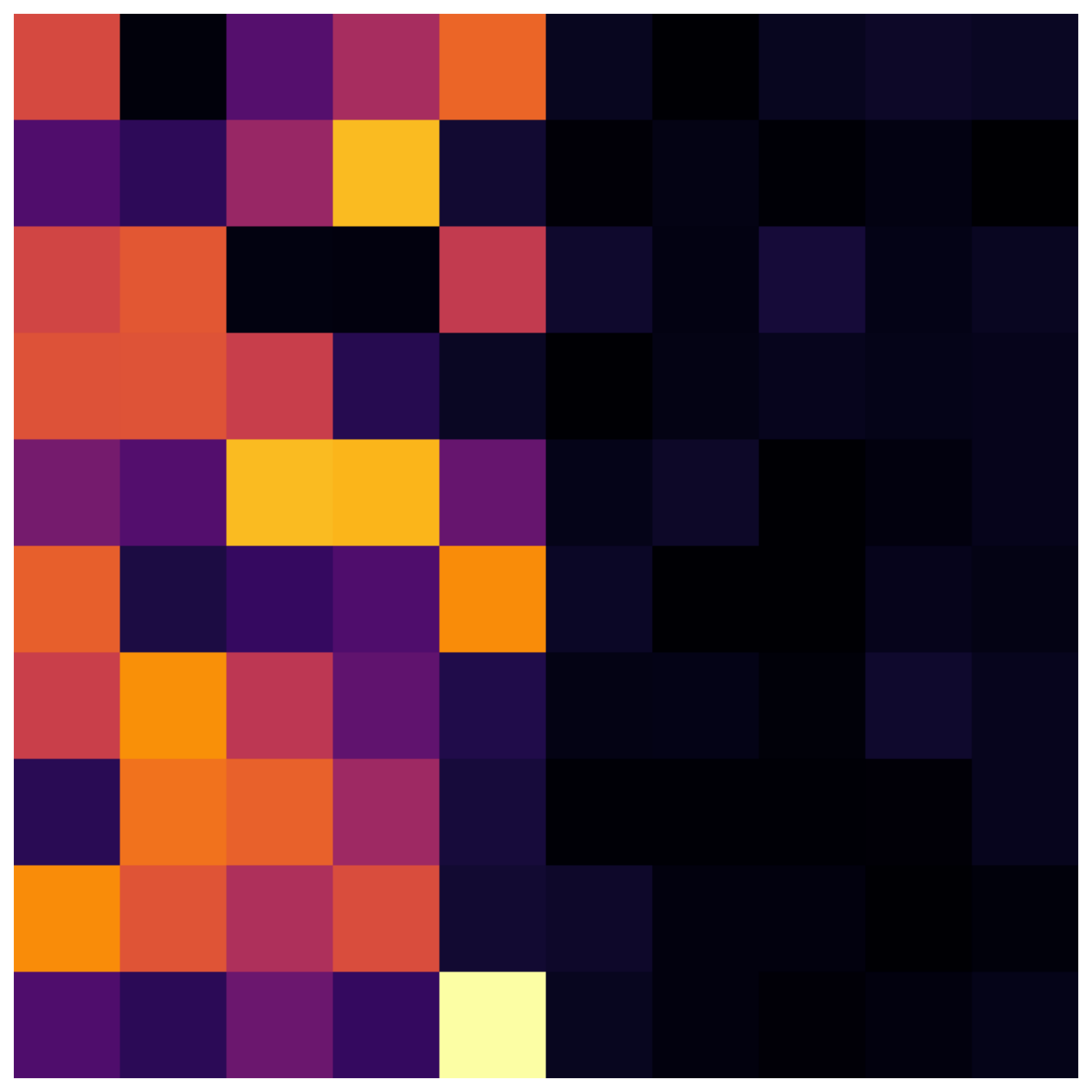}\\
    (A) FSAL features
    \end{minipage}\hfill%
    \begin{minipage}{0.48\linewidth}
    \centering
    \includegraphics[width=\linewidth]{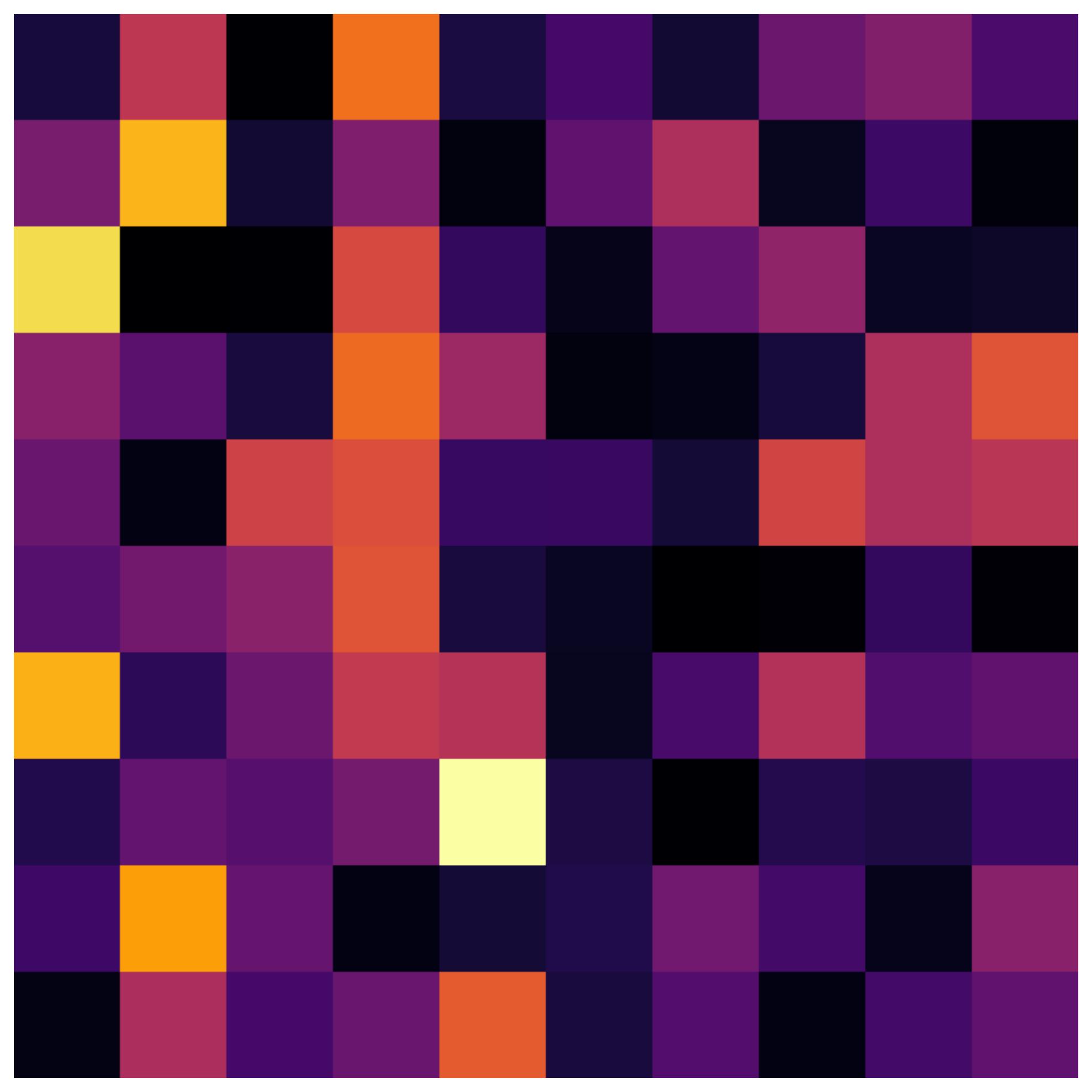}\\
    (B) FSL task learned features
    \end{minipage}
    \caption{Projecting data features into prototypical network embedding space ($WA$) for the
    linear toy problem. Values closer to zero are darker in colour. On the FSAL task, the model
    destroys information from the test attributes to remove ambiguity at training time.}
    \label{fig:toy_problem_weights}
\end{figure}
\paragraph{Linear prototypical network} Now, consider training a prototypical network on this data
with a linear embedding network, $g(\bx)~=~W\bx$. Within each episode, the prototypical network
computes the prototypes for the positive and negative examples,
\[
\bc_{j} = \frac{1}{k}\sum_{\bx_i \in S_j} g(\bx_i)
= \frac{1}{k}\sum_{\bx_i \in S_j} \sum_{l=1}^{d}z_{il}W\ba_{l},\textrm{ for }j \in \{0, 1\},
\]
where $S_j$ is the set of data points in the episode with label $j$, and $\ba_{l}$ is the
$l^{\text{th}}$ column of the matrix $A$. Further, the prototypical network likelihood is given
by,
\[p(y=0|\bx) = \frac{\exp\left\{-\Vert W\bx - \bc_0 \Vert^2_2\right\}}{\exp\left\{-\Vert W\bx -
\bc_0 \Vert^2_2\right\} + \exp\left\{-\Vert W\bx - \bc_1 \Vert^2_2\right\}}.\] The goal of the
prototypical network is thus to learn weights $W$ that lead to small distances between data points
in the same class and large distances otherwise. In the FSAL tasks, there is
an additional challenge in that class boundaries shift between episodes. The context (the choice of attribute entries) defining the
boundary is unknown and must be inferred from the episode. However, with few shots (small $k$) there
is ambiguity in the correct context --- with a high probability that several possible contexts
provide valid explanations for the observed data.

\paragraph{Fitting the prototypical network}
Notice that under our generative model, with $\bx = W\bz + \bzeta$ and for $j \in \{0, 1\}$ we have,
\begin{align*}
W\bx - \bc_j =\ & W A (\bz - \frac{1}{k}\sum_{\bz_i \in S_j} \bz_i) + 
\frac{1}{k}\sum_{i}W\bzeta_i +
W\bzeta.
\end{align*}
Notice that if $\rvv_j(\bz) = A (\bz - \frac{1}{k}\sum_{\bz_i \in S_j} \bz_i)~\in~\textrm{Ker}(W)$,
the kernel of $W$, then the entire first term is zero. Further, if $\bz \in S_j$ (the same class as
the prototype) then there is no contribution from the positive attribute features in this term.
Otherwise, this term is guaranteed to have some contribution from the positive attribute features.

Therefore, if $W$ projects to the linear space spanned by the positive attribute features then
$W\rvv_j(\bz)$ is zero when $\bz \in S_j$ and non-zero otherwise. This means that the model will be
able to solve the episode without contextual ambiguity. Then the optimal weights are those that
project to the set of features used in the training set---destroying all information about the test
attributes which would otherwise introduce ambiguity.

We observed this effect empirically in Figure~\ref{fig:toy_problem_weights}, where we have plotted
the matrix $\mathrm{abs}(WA)$. Each column of these plots represents a column of $A$ mapped to the
prototypical network's embedding space. The first 5 columns correspond to attributes used at
training time, and the remaining 5 to those used at test time.

In the FSAL task described above, as our analysis suggests, the learned prototypical feature
weights project out the features used at test time (the last 5 columns). As a result, the model
achieved 100\% training accuracy but only 51\% test accuracy (chance is 50\%).

We also compared against an equivalent problem set up that resembles the standard few-shot learning setting. In the FSL problem, the binary attribute strings may have only a single non-zero entry and each episode is a binary classification problem where the learner must distinguish between two classes. Now the vector $\bz$ is a one-hot encoding and the comparison to the prototypes occurs only over a single feature column of $A$, thus there is no benefit to projecting out the test features. As expected, the model we learned (Figure~\ref{fig:toy_problem_weights} B) is not forced to throw away test-time information and achieves 100\% training accuracy and 99\% test accuracy.

\paragraph{Settings for Figure~\ref{fig:toy_problem_weights}} We use 10 attributes, 5 of which are
used for training and 5 for testing. We use a uniformly random sampled $A\in\bbR^{30\times10}$ and
the prototypical network learns $W\in\bbR^{10\times30}$. We use additive Gaussian noise when
sampling data points with a standard deviation of 0.1. The models are trained with the Adam
optimizer using default settings over a total of 30000 random episodes, and evaluated on an
additional 1000 test episodes. We used $k=20$ to produce these plots, but found that the result was
consistent over different shot counts.
\end{document}